\documentclass[11pt]{article}
\usepackage[utf8]{inputenc}
\pdfoutput=1
\usepackage{setspace}
\usepackage{palatino}
\usepackage{graphicx}
\usepackage{float}
\usepackage{titling} 
\usepackage{multirow}
\usepackage{lscape}
\usepackage{amsmath}
\usepackage{amssymb}
\usepackage{amsfonts}
\usepackage[a4paper, total={6in, 9.5in}]{geometry}
\fontfamily{ppl}\selectfont 
\usepackage{eqparbox}
\usepackage{arydshln}
 \usepackage[figuresright]{rotating}
\usepackage[american]{babel}
\usepackage{longtable}
\usepackage{authblk}
\RequirePackage{doi}

\usepackage{bm}
\usepackage{caption}
\usepackage{subcaption}
\usepackage[textsize=small]{todonotes}
\usepackage[normalem]{ulem}

\newcommand{\appendixqqsection}[1]{\addtocounter{section}{1}
   \setcounter{table}{0}
   \setcounter{figure}{0}
   \setcounter{equation}{0}
   \setcounter{subsection}{0}
  \section*{Supplementary \Alph{section}: #1}
}

\newcommand\appendixqq{
   \setcounter{section}{0}
   \renewcommand\thesection{\Alph{section}}
   \renewcommand{\thesubsection}{\thesection. \arabic{subsection}}
   \renewcommand{\thesubsubsection}{\thesubsection .\arabic{subsubsection}}
   \renewcommand{\thefigure}{\thesection.\arabic{figure}}
   \renewcommand{\thetable}{\thesection.\arabic{table}}
}

\usepackage[natbibapa]{apacite}
\PassOptionsToPackage{hyperindex,breaklinks}{hyperref}
\usepackage{hyperref}
\usepackage{xcolor}
\definecolor{darkblue}{rgb}{0, 0, 0.5}
\hypersetup{colorlinks=true,citecolor=darkblue, linkcolor=darkblue, urlcolor=darkblue}

\usepackage[textsize=small]{todonotes}

\title{
Correspondence Analysis and PMI-Based Word Embeddings: A Comparative Study}

\author{Qianqian Qi$^*$\\
   {1. Department of Mathematics, \\Faculty of Sciences, Hangzhou Dianzi University, Hangzhou, China;}\\
   {2. Department of Methodology and Statistics, \\Faculty of Social Sciences, Utrecht University, Utrecht, The Netherlands}\\
   {*Corresponding author}\\
   \href{mailto:q.qi@hdu.edu.cn}{\texttt{q.qi@hdu.edu.cn}} 
\and Ayoub Bagheri\\
   {1. Department of Methodology and Statistics, \\Faculty of Social Sciences, Utrecht University, Utrecht, The Netherlands}\\
   \href{mailto:a.bagheri@uu.nl}{\texttt{a.bagheri@uu.nl}}
\and David J. Hessen\\
   {1. Department of Methodology and Statistics, \\Faculty of Social Sciences, Utrecht University, Utrecht, The Netherlands}\\
   \href{mailto:d.j.hessen@uu.nl}{\texttt{d.j.hessen@uu.nl}}
\and Peter G. M. van der Heijden\\
    {1. Department of Methodology and Statistics, \\Faculty of Social Sciences, Utrecht University, Utrecht, The Netherlands;}\\{2. Southampton Statistical Sciences Research Institute, \\University of Southampton, Highfield, Southampton, UK}\\
\href{mailto:p.g.m.vanderheijden@uu.nl}{\texttt{p.g.m.vanderheijden@uu.nl}}
    }
    
\date{}
\date{\vspace{-5ex}}
\begin{document}

{\setstretch{.8}
\maketitle
\begin{abstract}

Popular word embedding methods such as GloVe and Word2Vec are related to the factorization of the
pointwise mutual information (PMI) matrix. In this paper, we establish a formal connection between correspondence analysis (CA) and PMI-based word embedding methods. CA is a dimensionality reduction method that uses singular value decomposition (SVD), and we show that CA is mathematically close to the weighted factorization of the PMI matrix. We further introduce variants of CA for word-context matrices, namely CA applied after a square-root transformation (ROOT-CA) and after a fourth-root transformation (ROOTROOT-CA). We analyze the performance of these methods and examine how their success or failure is influenced by extreme values in the decomposed matrix. Although our primary focus is on traditional
static word embedding methods, we also include a comparison with a transformer-based encoder (BERT) to situate the results relative to contextual embeddings. Empirical evaluations across multiple corpora and word-similarity benchmarks show that ROOT-CA and ROOTROOT-CA perform slightly better overall than standard PMI-based methods and achieve results competitive with BERT.


\end{abstract}
}

\section{Introduction}

Word embeddings, i.e., dense and low dimensional word representations, are useful in various natural language processing (NLP) tasks \citep{jurafsky2023vector, MohammadiBaghmolaei2023Word, Johnson2024detailed, ZHANG2025125439, JADS4382025opinion, Ravenda2025supervised}. Three successful methods to derive such word representations are related to the factorization of the pointwise mutual information (PMI) matrix, an important matrix in NLP \citep{egleston2021statistical, bae2021keyword, ALQAHTANI2023103338, Toprak2025PMI}. The PMI matrix is a weighted version of the word-context co-occurrence matrix and measures how often two words, a target word and a context word, co-occur, compared with what we would expect if the two words were independent. The analysis of a positive PMI (PPMI) matrix, where all negative values in a PMI matrix are replaced with zero \citep{turney2010frequency, zhang2022word, ALQAHTANI2023103338, Toprak2025PMI}, generally leads to a better performance in semantic tasks \citep{bullinaria2007extracting}, and in most applications the PMI matrix is replaced with the PPMI matrix \citep{salle2016matrix}.

The first method, PPMI-SVD, decomposes the PPMI matrix with a singular value decomposition (SVD) \citep{levy2014neural, levy2015improving, stratos2015model, zhang2022word, kiyama2025analyzing}. The second one is GloVe \citep{pennington2014glove}. GloVe factorizes the logarithm of the word-context matrix with an adaptive gradient algorithm (AdaGrad) \citep{duchi2011adaptive}. According to \citet{shi2014linking, shazeer2016swivel},
GloVe is almost equivalent to factorizing a PMI matrix shifted by the logarithm of the sum of the elements of a word-context matrix. The third method is Word2Vec's skip-gram with negative sampling (SGNS) \citep{mikolov2013efficient, mikolov2013distributed}. SGNS uses a neural network model to generate word embeddings. \citet{levy2014neural} proved that SGNS implicitly factorizes a PMI matrix shifted by the logarithm of the number of negative samples in SGNS.

In this paper we study what correspondence analysis (CA) \citep{greenacre2017correspondence, beh2021introduction, Handan-Nader14102025, favcevicova2025correspondence} has to offer for the analysis of word-context co-occurrence matrices. CA is an exploratory statistical method that is often used for visualization of a low dimensional approximation of a matrix. It is close to the T-test (TTEST) weighting scheme \citep{curran2002improvements, curran2004distributional}, where standardized residuals are studied, as CA is based on the SVD of the matrix of standardized residuals.
In the context of document-term matrices, CA has been compared earlier with latent semantic analysis (LSA), where the document-term matrix is also decomposed with an SVD \citep{dumais1988using, deerwester1990indexing}. Although CA is similar to LSA, there is theoretical and empirical research showing that CA is preferable to LSA for text categorization and information retrieval \citep{Qi_Hessen_Deoskar_van_der_Heijden_2024, qi2024improving}. 

A document-term matrix has some similarity to a word-context matrix, as they both use counts, and, as mentioned above, the PMI matrix is a weighted version of the word-context matrix. In this paper, mathematically, we show that CA is close to a weighted factorization of the PMI matrix. We also propose a direct weighted factorization of the PMI matrix (PMI-GSVD). We further compare the performance of CA with the performance of PMI-based methods on word similarity tasks, and analyze their success and failure in terms of extreme values in the decomposed matrices.

In the context of CA, \citet{Nishisato2021}
point out, generally speaking, a two-way contingency table is prone to overdispersion, which may negatively affect the performance of CA \citep{beh2018correspondence, Nishisato2021}. To deal with this overdispersion, a fourth-root transformation is commonly used \citep{field1982practical, greenacre2009power, greenacre2010log}.
A word-context matrix can be taken as a two-way contingency table. Accordingly, in addition to applying CA to the raw word-context matrix, CA is also applied to its fourth-root transformation (ROOTROOT-CA). The square-root transformation, which helps stabilize the variance of Poisson-distributed counts, can also be beneficial for word–context matrices \citep{stratos2015model}; we therefore apply CA to the square-root transformation matrix (ROOT-CA) as well, an approach previously explored in biology \citep{hsu2023correspondence}. We further compare them with ROOT-CCA, proposed by \citet{stratos2015model}, which is similar to ROOT-CA. Importantly, ROOT-CA and ROOTROOT-CA represent CA variants that have not yet been explored in the NLP context.

Transformer-based encoders,
such as BERT, are currently popular approach in NLP \citep{kenton2019bert, Khan2025hybrid}. While transformer-based encoders are powerful for contextual embeddings, they often struggle with tasks which are related to single word embeddings. For example, \cite{aida2021comprehensive} shows PMI-based methods are better than BERT in the measuring semantic differences. This paper focuses on traditional PMI-based word embedding methods. To enhance the contribution of this paper, we also incorporate comparison of BERT with these traditional methods.

Considering the foregoing, this study focuses on word embeddings in NLP. First, we investigate the relationship between CA and PMI-based methods and compare their performance on word similarity tasks. Second, we introduce two variants of CA, namely ROOT-CA and ROOTROOT-CA, which have not previously been explored in the NLP literature. Third, we analyze the success and failure of these methods in terms of extreme values in the matrix before SVD, thereby providing insights into how to improve SVD-based methods. Additionally, we compare the performance of TTEST, PMI, PPMI, WPMI, ROOT-TTEST, ROOTROOT-TTEST, and STRATOS-TTEST, where SVD has not yet been applied.

In Section~\ref{S: CA}, CA, the three variants of CA, and the T-Test weighting scheme are introduced. The three PMI-based methods (PPMI-SVD, GloVe, and SGNS) and BERT are described in Section~\ref{S: Pmibasedmodel}. Theoretical relationships between CA and the PMI-based methods as well as BERT are shown in Section~\ref{S: Rela}, where PMI-GSVD is also introduced. In Section~\ref{S: datasets} we present three corpora to build word vectors and four word similarity datasets to evaluate word vectors. Section~\ref{S: expsetup} illustrates the setup of the empirical study using these three corpora. Section~\ref{S: preres} presents the results for these methods on word similarity tasks and analyzes the reason of these methods'performance in terms of extreme values. Section~\ref{S: discussion} and Section~\ref{S: conclusion} discuss and conclude this paper, respectively.

\section{Correspondence analysis}\label{S: CA}

In this section, first we describe correspondence analysis (CA) using a distance interpretation \citep{benzecri1973analyse, greenacre1987geometric}, which is a popular way to present CA. Then we present CA making use of an objective function, thus making the later comparison with PMI-based methods straightforward. Third, we present three variants of CA in word embedding. Finally, the T-Test weighting scheme \citep{curran2002improvements, curran2004distributional} is described, as it turns out to be remarkably similar to CA.

A word-context matrix is a matrix with counts, in which the rows and columns are labeled by terms. In each cell a count represents the number of times the row (target) word and the column (context) word co-occur in a text \citep{jurafsky2023vector}. Consider a word-context matrix denoted as $\bm{X}$ having $I$ rows ($i = 1, 2, \cdots, I$) and $J$ columns ($j = 1, 2, \cdots, J$), where the element for row $i$ and column $j$ is $x_{ij}$. The joint observed proportion is $p_{ij} = x_{ij}/x_{++}$, where "+" represents the sum over the corresponding elements and $x_{++} = \sum_{i}\sum_{j}x_{ij}$. The marginal proportions of target word $i$ and context word $j$ are $p_{i+} = \sum_jp_{ij}$ and $p_{+j} = \sum_ip_{ij}$, respectively.

\subsection{Introduction to CA}

CA is an exploratory method for the analysis of two-way contingency tables. It allows to study how the counts in the contingency table depart from statistical independence. Here we introduce CA in the context of the word-context matrix $\bm{X}$. In CA of the matrix $\bm{X}$, first the elements $x_{ij}$ are converted to joint observed proportions $p_{ij}$, and these are transformed into standardized residuals  \citep{greenacre2017correspondence} 
\begin{equation}
\label{staresi}
\frac{p_{ij}-p_{i+}p_{+j}}{\sqrt{p_{i+}p_{+j}}}.
\end{equation}
Then an SVD is applied to this matrix of standardized residuals, yielding
\begin{equation}
\label{SVDCA1}
\frac{p_{ij}-p_{i+}p_{+j}}{\sqrt{p_{i+}p_{+j}}} =  \sum_{k = 1}^{\text{min}\left(I-1, J-1\right)} \sigma_k u_{ik} v_{jk},
\end{equation}

\noindent where $\sigma_k$ is the $k$th singular value, with singular values in the decreasing order, and $[u_{1k}, u_{2k}, \cdots, u_{Ik}]^T$ and $[v_{1k}, v_{2k}, \cdots, v_{Jk}]^T$ are the $k$th left and right singular vectors, respectively. When $\bm{X}$ has full rank, the maximum dimensionality is $\text{min}\left(I-1, J-1\right)$, where the $''-1''$ is due to the subtraction of elements $p_{i+}p_{+j}$, that leads to a centering of the elements of $\bm{X}$ as $\sum_i (p_{ij} - p_{i+}p_{+j}) = 0 = \sum_j (p_{ij} - p_{i+}p_{+j}$). Multiplying the singular vectors consisting of elements $u_{ik}$ and $v_{jk}$ by $p_{i+}^{-\frac{1}{2}}$ and $p_{+j}^{-\frac{1}{2}}$, respectively, leads to
\begin{equation}
\label{SVDCA2}
\frac{p_{ij}}{p_{i+}p_{+j}} - 1 =  \sum_{k = 1}^{\text{min}\left(I-1, J-1\right)} \sigma_k \phi_{ik} \gamma_{jk},
\end{equation}
where $\phi_{ik} = p_{i+}^{-\frac{1}{2}}u_{ik}$ and $\gamma_{jk} = p_{+j}^{-\frac{1}{2}}v_{jk}$. Scores $\phi_{ik}, k = 1, 2, \cdots, K$ and $\gamma_{jk}, k = 1, 2, \cdots, K$ provide the standard coordinates of row point $i$ and column point $j$ in $K$-dimensional space, respectively, because of $\sum_ip_{i+}\phi_{ik} = \sum_jp_{+j}\gamma_{jk} = 0$ and $\sum_ip_{i+}\phi_{ik}^2 = \sum_jp_{+j}\gamma_{jk}^2 = 1$. Scores $\phi_{ik}\sigma_k, k = 1, 2, \cdots, K$ and $\gamma_{jk}\sigma_k, k = 1, 2, \cdots, K$ provide the principle coordinates of row point $i$ and column point $j$ in $K$-dimensional space, respectively. When $K < \text{min}(I-1,J-1)$, the Euclidean distances between these row (column) points approximate so-called $\chi^2$-distances between rows (columns) of $\bm{X}$. The squared $\chi^2$-distance between rows $i$ and $i'$ of $\bm{X}$ is
\begin{equation}
\label{chisquare}
\delta_{ii'}^2 = \sum_j\frac{\left(\frac{p_{ij}}{p_{i+}} - \frac{p_{i'j}}{p_{i'+}}\right)^2}{p_{+j}},
\end{equation}
and similarly for the chi-squared distance between columns $j$ and $j'$. 
Equation~(\ref{chisquare}) shows that the $\chi^2-$distance $\delta_{ii'}$ measures the difference between the $i$th vector of conditional proportions $p_{ij}/p_{i+}$ and the $i'$th vector of conditional proportions $p_{i'j}/p_{i'+}$, where more weight is given to the differences in elements $j$ if $p_{+j}$ is relatively smaller compared to other columns.

Although the use of Euclidean distance is standard in CA, \citet{qi2024improving} show that for information retrieval cosine similarity leads to the best performance among Euclidean distance, dot similarity, and cosine similarity. The superiority of cosine similarity also holds in the context of word embedding studies \citep{bullinaria2007extracting}. Therefore, in this paper we use cosine similarity to calculate the similarity of row points and of column points. It is worth noting that $p_{i+}^{-\frac{1}{2}}$ in $\phi_{ik} = p_{i+}^{-\frac{1}{2}}u_{ik}$ and $p_{+j}^{-\frac{1}{2}}$  in $\gamma_{jk} = p_{+j}^{-\frac{1}{2}}v_{jk}$ have no effects on the cosine similarity. Details are in Supplementary materials A. We coin scores $u_{ik}\sigma_k, k = 1, 2, \cdots, K$ and $v_{jk}\sigma_k, k = 1, 2, \cdots, K$ an alternative coordinates system for CA directly suited for cosine similarity. 

The so-called total inertia is
\begin{equation*}
\label{E: totalinertiap}
 \sum_i\sum_j\frac{\left(p_{ij} - p_{i+}p_{+j}\right)^2}{p_{i+}p_{+j}} = \sum_{k = 1}^{\text{min}\left(I-1, J-1\right)}\sigma_k^2.
\end{equation*}
This illustrates that CA decomposes the total inertia over  $\text{min}\left(I-1, J-1\right)$ dimensions. The total inertia equals the well-known Pearson $\chi^2$ statistic divided by $x_{++}$, so that the total inertia does not depend on the sample size $x_{++}$. The relative contribution of cell $(i, j)$ to the total inertia is calculated as $\frac{\left(p_{ij} - p_{i+}p_{+j}\right)^2}{p_{i+}p_{+j}} / \sum_i\sum_j\frac{\left(p_{ij} - p_{i+}p_{+j}\right)^2}{p_{i+}p_{+j}}$. The relative contribution of the $i$th row ($j$th column) to the $k$th dimension is calculated as $u_{ik}^2$ ($v_{jk}^2$).

\subsection{The objective function of CA}

To simplify the later comparison of CA with the other models, we present the objective function that is minimized in CA. The objective function is \citep[pp.~345-349]{greenacre1984theory}:
\begin{equation}
\label{objectiveCA}
\sum_{i,j}p_{i+}p_{+j}\left(\frac{p_{ij}}{p_{i+}p_{+j}} - 1 - \bm{e}_i^T\bm{o}_j\right)^2,
\end{equation}

\noindent where $\bm{e}_i$ and $\bm{o}_j$ are parameter vectors for target word $i$ and context word $j$, with respect to which the objective function is minimized.  The vectors have length $K \leq \text{min}\left(I-1, J-1\right)$. We call the part of the formula to be approximated, i.e.  $\left(p_{ij}/p_{i+}p_{+j} - 1\right)$, the fitting function and the weighting part $p_{i+}p_{+j}$ the weighting function. Thus, according to (\ref{objectiveCA}), CA can be viewed as a weighted matrix factorization  of $\left(p_{ij}/p_{i+}p_{+j} - 1\right)$ with weighting function $p_{i+}p_{+j}$.

The solution is found using the SVD as in Equation~(\ref{SVDCA2}). The $K$-dimensional approximation of $\left(p_{ij}/p_{i+}p_{+j} - 1\right)$ is 
\begin{equation*}
\label{capmiappro}
\frac{p_{ij}}{p_{i+}p_{+j}} - 1  
 \approx \sum_{k=1}^{K} \sigma_{k} \phi_{ik}\gamma_{jk}
 = \bm{e}_i^T \bm{o}_j.
\end{equation*}
The matrix $[\bm{e}_i^T\bm{o}_j]$ minimizes (\ref{objectiveCA}) amongst all matrices of rank $K$ in a weighted least-squares sense \citep{greenacre1984theory}. The parameter vectors $\bm{e}_i$ and $\bm{o}_j$ can be represented, for example, as
\begin{equation*}
\label{casolutionword}
\bm{e}_i = [\phi_{i1}\sigma_1, \phi_{i2}\sigma_2, \cdots, \phi_{iK}\sigma_K]^T
\end{equation*}
and 
\begin{equation*}
\label{casolutioncontext}
\bm{o}_j = [\gamma_{j1}, \gamma_{j2}, \cdots, \gamma_{jK}]^T
\end{equation*}
As described above, this representation $\bm{e}_i$ of target word $i$ has the advantage that the $\chi^2$-distance between target words $i$ and $i'$ in the original matrix is approximated by the Euclidean distance between $\bm{e}_i$ and $\bm{e}_{i'}$.

The parameters $\bm{e}_i$ can be adjusted by a singular value weighting exponent $p$, i.e., $\bm{e}_i = [\phi_{i1}\sigma_1^p, \phi_{i2}\sigma_2^p, \cdots, \phi_{iK}\sigma_K^p]^T$. Correspondingly, the alternative coordinate for the adjusted row $i$ by a singular value weighting exponent is $ [u_{i1}\sigma_1^p, u_{i2}\sigma_2^p, \cdots, u_{iK}\sigma_K^p]^T$.

\subsection{Three variants of CA for word embeddings}\label{Sub: reca}

We present three variants of CA: ROOT-CCA, ROOT-CA, and ROOTROOT-CA, where ROOT-CA and ROOTROOT-CA are new to NLP and ROOT-CCA was introduced to NLP by \citet{stratos2015model}. According to \citet{stratos2015model}, word counts can be naturally modeled as Poisson variables. The square-root transformation of a Poisson variable leads to stabilization of the variance \citep{bartlett1936square, stratos2015model}. \citet{stratos2015model} proposed to combine canonical correlation analysis (CCA) with the square-root transformation of the word-context matrix. Even though CA of a contingency table is equivalent to CCA of the data in the form of an indicator matrix, we call the proposal by \citet{stratos2015model} ROOT-CCA, to distinguish it from ROOT-CA, discussed later. 

\paragraph{ROOT-CCA}
In ROOT-CCA, an SVD is performed on the matrix whose typical element is the square root of $x_{ij}/\sqrt{x_{i+}x_{+j}} = p_{ij}/\sqrt{p_{i+}p_{+j}}$, that is
\begin{equation}\label{E: ccaroot}
    \sqrt{\frac{p_{ij}}{\sqrt{p_{i+}{p_{+j}}}}} =  \sum_{k = 1}^{\text{min}(I, J)} \sigma_k u_{ik} v_{jk}.
\end{equation}
The reason that \citet{stratos2015model} ignore $p_{i+}p_{+j}$ in $p_{ij}-p_{i+}p_{+j}$ (compare Equation (\ref{SVDCA1})) is that they believe that, when the sample size $x_{++}$ is large, the first part $p_{ij}/\sqrt{(p_{i+}p_{+j})}$ in $(p_{ij}-p_{i+}p_{+j})/\sqrt{(p_{i+}p_{+j})}$ dominates the expression. 

\paragraph{ROOT-CA}

Inspired by \citet{stratos2015model}, we present CA of the square-root transformation of the word-context matrix  (ROOT-CA) \citep{bartlett1936square, hsu2023correspondence}. ROOT-CA differs from ROOT-CCA in the following way. In the ROOT-CA, first we create a square-root transformation of the word-context matrix with elements $\sqrt{x_{ij}}$, and then we perform CA on this matrix. Let $p_{ij}^* = \frac{\sqrt{x_{ij}}}{\sum_{ij}\sqrt{x_{ij}}} = \frac{\sqrt{p_{ij}}}{\sum_{ij}\sqrt{p_{ij}}}$. Then ROOT-CA provides the decomposition 
\begin{equation}\label{E: rootca}
  \frac{p_{ij}^*-p_{i+}^*p_{+j}^*}{\sqrt{p_{i+}^*p_{+j}^*}}  =  \sum_{k = 1}^{\text{min}(I-1, J-1)} \sigma_k u_{ik} v_{jk}.
\end{equation}

\paragraph{ROOTROOT-CA}\label{Subsubsub: rootrootca}

According to \citet{stratos2015model}, word counts can be naturally modeled as Poisson variables. In the Poisson distribution the mean and variance are identical. The phenomenon of the data having greater variability than expected based on a statistical model is called overdispersion \citep{agresti2007categorical}. In the context of CA, \citet{Nishisato2021} point out, generally speaking, a two-way contingency table is prone to overdispersion. Overdispersion may negatively affect the performance of CA \citep{beh2018correspondence, Nishisato2021}.

\citet{greenacre2009power, greenacre2010log}, referring to \citet{field1982practical}, points out that in ecology abundance data is almost always highly over-dispersed and a particular school of ecologists routinely applies a fourth-root transformation before proceeding with the statistical analysis. Therefore we also study the effect of a root-root transformation before performing CA. We call it ROOTROOT-CA. That is, ROOTROOT-CA is a CA on the matrix with typical element $\sqrt{\sqrt{x_{ij}}}$ \citep{field1982practical}. Suppose $p_{ij}^{**} = \frac{\sqrt{\sqrt{x_{ij}}}}{\sum_{ij}\sqrt{\sqrt{x_{ij}}}} = \frac{\sqrt{\sqrt{p_{ij}}}}{\sum_{ij}\sqrt{\sqrt{p_{ij}}}}$. Then, we have
\begin{equation}\label{E: rootrootca}
  \frac{p_{ij}^{**}-p_{i+}^{**}p_{+j}^{**}}{\sqrt{p_{i+}^{**}p_{+j}^{**}}}  =  \sum_{k = 1}^{\text{min}(I-1, J-1)} \sigma_k u_{ik} v_{jk},
\end{equation}
Thus ROOT-CA and ROOTROOT-CA  are pre-transformations of the  elements $x_{ij}$ of the original matrix by $\sqrt{x_{ij}}$ and $\sqrt{\sqrt{x_{ij}}}$, respectively. CA is performed on the transformed matrix.

\subsection{T-Test}\label{Subsec: TTest}

The T-Test (TTEST) weighting scheme, described by \citet{curran2002improvements} and \citet{curran2004distributional}, focuses on the matrix of standardized residuals, see Equation~(\ref{staresi}). Thus it is remarkably similar to CA, where the matrix of standardized residuals is decomposed. 
For a comparison between CA and TTEST weighting in word similarity tasks, as we will carry out below, the question is whether the performance is better on the matrix of standardized residuals, or on a low dimensional representation of this matrix provided by CA.

Inspired by Section~\ref{Sub: reca}, we also explore the performances of the matrix STRATOS-TTEST with typical element $\sqrt{p_{ij} / \sqrt{p_{i+}{p_{+j}}}}$ (compare Equation~(\ref{E: ccaroot})), the matrix ROOT-TTEST with typical element $\left(p_{ij}^*-p_{i+}^*p_{+j}^*\right) / \sqrt{p_{i+}^*p_{+j}^*}$ (compare Equation~(\ref{E: rootca})), and the matrix ROOTROOT-TTEST with typical element $\left(p_{ij}^{**}-p_{i+}^{**}p_{+j}^{**}\right) / \sqrt{p_{i+}^{**}p_{+j}^{**}}$ (compare Equation~(\ref{E: rootrootca})).

\section{PMI-based word embedding methods and BERT}\label{S: Pmibasedmodel}

\subsection{PMI-SVD and PPMI-SVD}\label{Subsec: pmi}

Pointwise mutual information (PMI) is an important concept in NLP. The PMI between a target word $i$ and a context word $j$ is defined as \citep{egleston2021statistical, bae2021keyword, ALQAHTANI2023103338, jurafsky2023vector, Toprak2025PMI}:
\begin{equation*}
\label{Contingencyratio}
\text{PMI}(i,j) = \text{log }\frac{p_{ij}}{p_{i+}p_{+j}}
\end{equation*}
i.e. the log of the contingency ratios \citep{greenacre2009power, greenacre2017correspondence}, also known as Pearson ratios \citep{Leo1996, beh2021introduction}, ${p_{ij}}/\left(p_{i+}p_{+j}\right)$. If $p_{ij} = 0$, then $\text{PMI}(i, j) = \text{log } 0 = -\infty$, and it is usual to set $\text{PMI}(i, j) = 0$ in this situation.

A common approach is to factorize the
PMI matrix using SVD, which we call PMI-SVD. Thus the objective function is
\begin{equation}
\label{objectivePMISVD}
\sum_{i,j}\left(\text{log }\frac{p_{ij}}{p_{i+}p_{+j}} - \bm{e}_i^T\bm{o}_j\right)^2.
\end{equation}
In terms of a weighted matrix factorization, PMI-SVD is the matrix factorization of the PMI matrix with the weighting function $1$. The solution is provided directly via SVD. An SVD applied to the PMI matrix with elements $\text{log}\left(p_{ij}/\left(p_{i+}p_{+j}\right)\right)$ yields

\begin{equation*}
\label{SVDPMI1}
\text{log }\frac{p_{ij}}{p_{i+}p_{+j}} =  \sum_{k = 1}^{\text{min} (I, J)}  \sigma_k u_{ik} v_{jk},
\end{equation*}

\noindent where $\text{min}(I, J)$ is the rank of the PMI matrix. The $K$-dimensional approximation of $\text{log}\left(p_{ij}/\left(p_{i+}p_{+j}\right)\right)$ is 
\begin{equation*}
\label{pmiappro}
\text{log }\frac{p_{ij}}{p_{i+}p_{+j}}
 \approx \sum_{k=1}^{K} \sigma_{k} u_{ik}v_{jk}
 = \bm{e}_i^T\bm{o}_j
\end{equation*}
where the matrix with elements $\bm{e}_i^T\bm{o}_j$ minimizes (\ref{objectivePMISVD}) amongst all matrices of rank $K$ in a least squares sense, where $K \leq \text{min} (I, J)$. Both CA and PMI-SVD are dimensionality reduction techniques making use of SVD.

The parameters $\bm{e}_i$ and $\bm{o}_j$ can be represented as
\begin{equation*}
\label{pmisvdsolutionword}
\bm{e}_i = \left[u_{i1}\sigma_1, u_{i2} \sigma_2, \cdots, u_{iK}\sigma_K\right]^T
\end{equation*}
and 
\begin{equation*}
\label{pmisvdsolutioncontext}
\bm{o}_j = \left[v_{j1}, v_{j2}, \cdots, v_{jK}\right]^T.
\end{equation*}
Thus the Euclidean distance between target words $i$ and $i'$ in the original matrix is approximated by the Euclidean distance between $\bm{e}_i$ and $\bm{e}_{i'}$. In practice, one regularly sees that the parameters $\bm{e}_i$ are adjusted by an exponent $p$ used for weighting the singular values, i.e., $\bm{e}_i = \left[u_{i1}\sigma_1^p, u_{i2}\sigma_2^p, \cdots, u_{iK}\sigma_K^p\right]^T$, where $p$ is usually set to 0 or 0.5 \citep{levy2014neural, levy2015improving, stratos2015model}.

It is worth noting that the elements in the PMI matrix, where word-context pairs that co-occur rarely are negative, but word-context pairs that never co-occur are set to 0 \citep{levy2014neural}, are not monotonic transformations of observed counts divided by counts under independence. For this reason an alternative is proposed, namely the  positive PMI matrix, abbreviated as PPMI matrix. In the PPMI matrix all negative values are set to 0:
\begin{equation*}
\label{Contingencyratioppmi}
\text{PPMI}(i,j) = \text{max}\left(\text{PMI}(i,j), 0\right)
\end{equation*}
In most applications, one makes use of the PPMI matrix instead of the PMI matrix \citep{salle2016matrix}. We call the factorization of the PPMI matrix using SVD PPMI-SVD \citep{zhang2022word}.

\subsection{GloVe}\label{Subsec: Glove}

The GloVe objective function to be minimized is  \citep{pennington2014glove}:
\begin{equation}
\label{objectiveGlove}
\sum_{i,j} f(x_{ij})\left(\text{log }x_{ij} - b_i - s_j-\bm{e}_i^T\bm{o}_j\right)^2
\end{equation}
where 
\begin{equation*}
f(x_{ij})=\begin{cases}
          \left(x_{ij}/x_{\text{max}}\right)^{\alpha} &\text{if} \, x_{ij} < x_{\text{max}} \\
          1 & \, \text{otherwise} \\
     \end{cases}
\end{equation*}
In addition to parameter vectors $\bm{e}_i$ and $\bm{o}_j$, the scalar parameter terms $b_i$ and $s_j$ are referred to as $bias$ of target word $i$ and context word $j$, respectively.
\citet{pennington2014glove} train the GloVe model using an adaptive gradient algorithm (AdaGrad)
\citep{duchi2011adaptive}. This algorithm trains only on the non-zero elements of a word-context matrix, as $f(0) = 0$, which avoids the appearance of the undefined $\text{log }0$ in Equation~(\ref{objectiveGlove}).

In the original proposal of GloVe \citep{pennington2014glove}, $b_i = \text{log }x_{i+}$ and then, due to the symmetric role of target word and context word, $s_j = \text{log }x_{+j}$. \citet{shi2014linking} and \citet{shazeer2016swivel} show that the bias terms $b_i$ and $s_j$ are highly correlated with $\text{log }x_{i+}$ and $\text{log }x_{+j}$, respectively, in GloVe model training. This means that the GloVe model minimizes a weighted least squares loss function with the weighting function $f(x_{ij})$ and approximate fitting function $\text{log}\,x_{ij} - \text{log}\,x_{i+} - \text{log}\,x_{+j} = \text{log}\left(x_{ij}x_{++}/\left(x_{i+}x_{+j}\right)\right) - \text{log}\,x_{++} = \text{log}\left(p_{ij}/\left(p_{i+}p_{+j}\right)\right) - \text{log}\,x_{++}$:
\begin{equation}
\label{objectiveGlovealter}
\sum_{i,j} f(x_{ij})\left(\text{log }\frac{p_{ij}}{p_{i+}p_{+j}} - \text{log }x_{++}-\bm{e}_i^T\bm{o}_j\right)^2
\end{equation}

\subsection{Skip-gram with negative sampling}\label{Subsec: SGNS}

SGNS stands for skip-gram
with negative sampling of word2vec embeddings \citep{mikolov2013efficient, mikolov2013distributed}. The algorithms used in SGNS are stochastic gradient descent and backpropagation \citep{rumelhart1986learning, rong2014word2vec}. SGNS trains word embeddings on every word of the corpus one by one.

\citet{levy2014neural} showed that SGNS implicitly factorizes a PMI matrix shifted by $\text{log }n$:
\begin{equation*}
\label{objectivelocalSGNSparsolueo}
\text{log }\frac{p_{ij}}{p_{i+}p_{+j}} - \text{log }n \approx \bm{e}_i^T\bm{o}_j
\end{equation*}
where $n$ is the number of negative samples. According to \citet{levy2014neural} and \citet{shazeer2016swivel}, the objective function of SGNS is approximately a minimization of the difference between $\bm{e}_i^T\bm{o}_j$ and $\text{log }\left(p_{ij}/\left(p_{i+}p_{+j}\right) - \text{log }n\right)$, tempered by
a monotonically increasing weighting function of the observed
co-occurrence count $x_{ij}$, that we denote by $g(x_{ij})$:

\begin{equation}
\label{objectiveword2vecalter}
\sum_{i,j} g(x_{ij})\left(\text{log }\frac{p_{ij}}{p_{i+}p_{+j}} - \text{log }n -\bm{e}_i^T\bm{o}_j\right)^2
\end{equation}
\noindent This demonstrates that the approximation of the SGNS objective in (\ref{objectiveword2vecalter}) differs from that of GloVe in (\ref{objectiveGlovealter}) in the use of $n$ instead of $x_{++}$ and $g(x_{ij})$ instead of $f(x_{ij})$. 

The above three methods PMI-SVD/PPMI-SVD, GloVe, and SGNS are related with PMI matrix. These three methods are static word embedding methods. I.e., each word has a single word embedding. This paper focuses on the static word embeddings. The transformer-based encoders with contextual word embeddings, such as BERT, are popular in NLP. In contextual word embeddings method, each word usually has more than a single word embedding. To enhance the contribution of this paper, we also compare BERT with these traditional static word embeddings methods. Next, we introduces BERT.

\subsection{BERT}

BERT stands for Bidirectional Encoder Representations from
Transformers \citep{kenton2019bert}. BERT’s model architecture is a multi-layer bidirectional transformer encoder and can obtain contextual word embeddings, i.e., a word in different contexts has different word embeddings. BERT includes two steps: pre-training and fine-tuning. BERT pre-trains on unlabeled data over two unsupervised tasks: masked language model (MLM) and next sentence prediction (NSP). Fine-tuning step is especially for downstream tasks using labeled data from the downstream tasks.

\section{Relationships of CA to PMI-based models as well as BERT}\label{S: Rela}

\subsection{CA and PMI-SVD / PPMI-SVD}\label{Subsec: CAPMI}

In this section, we discuss PMI-SVD and PPMI-SVD together, as PMI and PPMI are the same except that in PPMI all negative values of PMI are set to 0.

CA is closely related to PMI-SVD. This becomes clear by comparing $\left(p_{ij}/\left(p_{i+}p_{+j}\right)-1\right)$ in (\ref{objectiveCA}) with $\text{PMI}(i, j) = \text{log}\left(p_{ij}/\left(p_{i+}p_{+j}\right)\right)$ in (\ref{objectivePMISVD}). The relation lies in a Taylor expansion of $\text{log}\left(p_{ij}/\left(p_{i+}p_{+j}\right)\right)$, namely that, if $x$ is small, $\text{log}(1+x) \approx x$ \citep{van1989combined}. Substituting $x$ with $p_{ij}/ (p_{i+}p_{+j})$ – 1 leads to:
\begin{equation}\label{approxcapmi}
\text{log }\frac{p_{ij}}{p_{i+}p_{+j}} \approx
 \frac{p_{ij}}{p_{i+}p_{+j}} - 1 
\end{equation}

\noindent This illustrates that if $\left(p_{ij}/p_{i+}p_{+j} - 1\right)$ is small, the objective function of CA approximates
\begin{equation}
\label{objectiveCAalter}
\sum_{i,j}p_{i+}p_{+j}\left(\text{log }\frac{p_{ij}}{p_{i+}p_{+j}} - \bm{e}_i^T\bm{o}_j\right)^2.
\end{equation}
From Equation~(\ref{objectiveCAalter}) it follows that CA is approximately a weighted matrix factorization of $\text{log }\left(p_{ij}/\left(p_{i+}p_{+j}\right)\right)$ with weighting function $p_{i+}p_{+j}$. The Equation (\ref{approxcapmi}) can also be obtained by the Box-Cox transformation of the contingency ratios, for example, \citet{greenacre2009power} and \citet{beh2024correspondence}, and we refer to their work for more details.

Comparing Equation~(\ref{objectiveCAalter}) with Equation~(\ref{objectivePMISVD}), both CA and PMI-SVD can be taken as weighted least squares methods having approximately the same fitting functions, namely $\left(p_{ij}/p_{i+}p_{+j} - 1\right)$ for CA and  $\text{log}\left(p_{ij}/\left(p_{i+}p_{+j}\right)\right)$ for PMI-SVD. Both make use of an SVD. 

However, they use different weighting functions, namely $p_{i+}p_{+j}$ in CA and $1$ in PMI-SVD. 
It has been argued that equally weighting errors in the objective function, as is the case in PMI-SVD, is not a good approach \citep{salle2016matrix, salle2022understanding}. For example, \citet{salle2022understanding} presented the reliability principle, that the objective function should have a weight on the reconstruction error that is a monotonically increasing function of the marginal frequencies of word and of context.
On the other hand, CA, unlike PMI-SVD, weights errors in the objective function with a weighting function equal to the product of the marginal proportions of word and context \citep{greenacre1984theory, greenacre2017correspondence, beh2021introduction}.

\subsubsection{PMI-GSVD}
\noindent The weighting function of PMI-SVD is 1 while in the approximate version of CA it is $p_{i+}p_{+j}$. Therefore, we also investigate the performance of a weighted factorization of the PMI matrix, where $p_{i+}p_{+j}$ is the weighting function:
\begin{equation*}
\label{objectiveGPMISVD}
\sum_{i,j}p_{i+}p_{+j}\left(\text{log }\frac{p_{ij}}{p_{i+}p_{+j}} - \bm{e}_i^T\bm{o}_j\right)^2.
\end{equation*}
Similar with CA, we use generalized SVD (GSVD) to find the optimum of the objective function (PMI-GSVD). That is, an SVD is applied as follows:
\begin{equation*}
\label{GSVDPMI1}
\sqrt{p_{i+}p_{+j}}\,\text{log }\frac{p_{ij}}{p_{i+}p_{+j}} =  \sum_{k = 1}^{\text{min} (I, J)}  \sigma_k u_{ik} v_{jk},
\end{equation*}
We call the matrix with typical element $\sqrt{p_{i+}p_{+j}}\,\text{log }\frac{p_{ij}}{p_{i+}p_{+j}}$ the WPMI matrix, also known as the modified log-likelihood ratio residual \citep{beh2024correspondence}.

\subsection{CA and GloVe}\label{Subsec: CAGlove}

Both CA and GloVe are weighted least squares methods. The weighting function in GloVe is $f(x_{ij})$, which is defined uniquely for each element of the word-context matrix, while  the weighting function $p_{i+}p_{+j}$ in CA is defined by the row and column margins. 

In the approximate fitting function of GloVe, $\text{log}\left(p_{ij}/\left(p_{i+}p_{+j}\right)\right) - \text{log }x_{++}$, the term  $\text{log }x_{++}$ can be considered as a shift of $\text{log}\left(p_{ij}/\left(p_{i+}p_{+j}\right)\right)$. And as we showed in Section~\ref{Subsec: CAPMI}, the fitting function of CA is approximately $\text{log}\left(p_{ij}/\left(p_{i+}p_{+j}\right)\right)$ when $p_{ij}$ is close to $p_{i+}p_{+j}$. Thus, from a comparison of the objective functions of CA and GloVe, it is natural to expect that these two methods will yield similar results if $\left(p_{ij}/p_{i+}p_{+j} - 1\right)$ is small.

In comparing the algorithms of these two methods, we find that CA uses SVD  while GloVe uses AdaGrad. These two algorithms have their own advantages and disadvantages. On the one hand, the AdaGrad algorithm trains  GloVe only on the nonzero elements of word-context matrix, one by one, while in CA the SVD decomposes the entire word-context matrix in full in one step. On the other hand, the SVD always finds the global minimum while the AdaGrad algorithm cannot guarantee the global minimum.

\subsection{CA and SGNS}\label{Subsec: CASGNS}

By comparing Equations~(\ref{objectiveword2vecalter}) and (\ref{objectiveCAalter}), both the approximation of CA and of SGNS are found by weighted least squares methods. The weighting function in SGNS is $g(x_{ij})$, which is defined for each element of word-context matrix where frequent word-context pairs pay more for deviations than infrequent ones \citep{levy2014neural}, while the weighting function  in CA is defined by the row and column margins, i.e. $p_{i+}p_{+j}$.

In the fitting function of the approximation of SGNS, $\text{log}\left(p_{ij}/\left(p_{i+}p_{+j}\right)\right) - \text{log }n$, the term $\text{log }n$ can be considered as a shift of $\text{log}\left(p_{ij}/\left(p_{i+}p_{+j}\right)\right)$. As shown in Section~\ref{Subsec: CAPMI}, the approximate fitting function in CA is $\text{log}\left(p_{ij}/\left(p_{i+}p_{+j}\right)\right)$. Thus, considering the objective function view, both the approximation of CA and of SGNS make use of the PMI matrix.

Although the approximate objective function of SGNS is similar to that of CA, the training processing for SGNS is different from that of CA. SGNS trains word embeddings on the words of a corpus, one by one, to maximize the probabilities of target words and context words co-occurrence, and to minimize the probabilities between target words and randomly sampled words, by updating the vectors of target words and context words.  
In contrast, CA first counts all co-occurrences in the corpus and then performs SVD on the matrix of standardized residuals to obtain the vectors of target words and context words at once.

\subsection{CA and BERT}

CA and BERT have different objective functions. BERT pre-train two tasks. One is MLM, which aims to predict the masked tokens. The other is NSP, which predicts whether the second sentence is followed by the first sentence. In contrast, CA minimizes the weighted least-square function, involving decomposing the matrix of standardized residuals via SVD. BERT is to predict using self-supervised learning, while CA focuses on unsupervised dimensionality reduction by explaining inertia in two categorical variables of word-context matrix.

The number of parameters in BERT is huge, where BERT$_{\text{BASE}}$ is around 110 million and BERT$_{\text{LARGE}}$ is around 340 million \citep{kenton2019bert}. Thus BERT needs large corpus and spends a lot of time to train. Therefore, BERT may not be suitable for low resources language and low computing power \citep{aida2021comprehensive}. In contrast, the number of parameters in CA is much less and thus requires lower computing power and lower resources language.

\section{Three corpora and four word similarity datasets}\label{S: datasets}

All static word embeddings methods are trained on three corpora: Text8 \citep{text8}, British National Corpus (BNC) \citep{BNC}, and Wikipedia from May, 2024 (Wiki052024) \citep{Wiki052024}, respectively. For BERT, we use pre-trained BERT$_{\text{BASE}}$ and use fine-tuned BERT$_{\text{BASE}}$ which fine-tunes the pre-trained BERT$_{\text{BASE}}$ on Wiki052024 (Details are in Section \ref{sub: setupbert}). 

Text8 is a widely used corpus in NLP \citep{xin2018batch, roesler2019evaluation, podkorytov2020effects, guo2021document}.
It includes more than 17 million words from Wikipedia \citep{peng2017automatic} and only consists of lowercase English characters and spaces. Words that appeared less than 100 times in the corpus are ignored, resulting in a vocabulary of 11,815 terms.

BNC is from a representative
variety of sources and is widely used \citep{raphael2023gendered, samuel-etal-2023-trained}. Data cited herein have been extracted from the British National Corpus, distributed by the University of Oxford on behalf of the BNC Consortium. We remove English punctuation and numbers and set words in lowercase form. Words that appeared less
than 500 times in the corpus are ignored, resulting in a vocabulary of 11,332 terms.

Wikipedia is a standard corpus used to create word embeddings \citep{mikolov2013efficient, mikolov2013distributed, pennington2014glove, levy2015improving, stratos2015model, kenton2019bert}. Wiki052024 is from the "enwiki-20240501-pages-articles-multistream.xml.bz2" which is downloaded from \cite{Wiki052024}. We use wikiextractor to extract and clean this corpus \citep{Wikiextractor2015}. Following BNC, we remove English punctuation and numbers and set words in lowercase form. This corpus has 144,581,193 sentences and 16,362,680,184 words, and has the vocabulary of 10,582,162. Due to the restriction of computing power, we are taking a smaller sample of the Wiki052024 by limiting the frequency threshold.  That is, we only choose vocabulary where the frequency is equal to or more than 10,000 times, which results in a vocabulary of 15,135 words.

Following previous studies \citep{levy2015improving, pakzad2021word}, we evaluate each word embeddings method on word similarity tasks using the Spearman’s
correlation coefficient $\rho$. We use four popular word similarity datasets: WordSim353 \citep{finkelstein2002placing},
MEN  \citep{bruni2012distributional}, Mechanical Turk  \citep{radinsky2011word}, and SimLex-999 \citep{hill-etal-2015-simlex}. All these datasets consist of
word pairs together with human-assigned similarity scores. For example, in WordSim353, where scores range from 0 (least similar) to 10 (most similar), one word pair is (tiger, cat) with human assigned similarity score 7.35.
Out-of-vocabulary words are removed from
all test sets. I.e.,  if either tiger or cat doesn't occur in the vocabularies of the 11,815 terms created by Text8 corpus, we delete (tiger, cat). Thus for evaluating the different word embedding methods in Text8 277 word pairs with scores are kept in WordSim353 instead of the original 353 word pairs. Table~\ref{wordsimevaluation} provides the number of word pairs used by the datasets in Text8, BNC, and Wiki052024. 

\begin{table}[htbp]
\centering  
    \caption{Datasets for word similarity evaluation.} 
\label{wordsimevaluation}
\begin{tabular}{lrrrr}    
\hline
Dataset& Word pairs& Word pairs  in Text8 & Word pairs  in BNC & Word pairs in Wiki052024
\\
\hline  
WordSim353& 353&277&276 & 294\\ 
MEN& 3000 &1544&1925 & 2159\\  
Turk& 287 & 221& 197 & 256\\  
SimLex-999&999&726&847 & 810\\  
\hline 
\end{tabular}  
\end{table}

After calculating the solutions for CA, PMI-SVD, PPMI-SVD, PMI-GSVD, ROOT-CA, ROOTROOT-CA, ROOT-CCA, GloVe, SGNS, and BERT, we obtain the word embeddings. We calculate the cosine similarity for each word pair in each word similarity dataset. For example, for WordSim353 using Text8, we obtain 277 cosine similarities. The Spearman’s correlation coefficient $\rho$ \citep{hollander2013nonparametric} between these similarities and the human similarity scores is calculated to evaluate these word embedding methods. Larger values are better. 
 
\section{Study setup}\label{S: expsetup}
 
\subsection{SVD-based methods}\label{Sec: expsetupsvd}

CA, PMI-SVD, PPMI-SVD, PMI-GSVD, ROOT-CA, ROOTROOT-CA, and ROOT-CCA are SVD-based dimensionality reduction methods. First, we create a word-context matrix
of size 11,815$\times$11,815, 11,332$\times$11,332, and 15,135$\times$15,135 based on Text8, BNC, and Wiki052024, respectively. We use a window of size 2, i.e., two words to each side of the target word. A context word one token and two tokens away will be counted as $1/1$ and $1/2$ of an occurrence, respectively. Then we perform SVD on the related matrices. We use the svd function from scipy.linalg in Python to calculate the SVD of a matrix, and obtain singular values $\sigma_k$, left singular vectors $u_{ik}$, and right singular vectors $v_{jk}$. We obtain the word embeddings as $\bm{e}_i = \left[u_{i1}\sigma_1^p, u_{i2}\sigma_2^p, \cdots, u_{ik}\sigma_k^p\right]^T$.

The choices of the exponent weighting $p$ and number of dimensions $k$ are important for SVD-based methods. In the context of PPMI-SVD and ROOT-CCA $p$ is regularly set to $p = 0$ or $p = 0.5$ \citep{levy2014neural, levy2015improving, stratos2015model}.  For $p = 0$, we have the standard coordinates with $\bm{U}^T\bm{U} = \bm{V}^T\bm{V} = \bm{I}$. For $p = 0.5$, we have $\bm{A}_k = \bm{U}_k\bm{\Sigma}_k\bm{V}_k^T = (\bm{U}_k\bm{\Sigma}_k^{1/2})(\bm{V}_k\bm{\Sigma}_k^{1/2})^T$. That is, the target words $\bm{U}_k\bm{\Sigma}_k^{1/2}$ and context words $\bm{V}_k\bm{\Sigma}_k^{1/2}$ reconstruct the decomposed matrix $\bm{A}_k$. The three created word-context matrices based on Text8, BNC, and Wiki052024 are symmetric, so the matrices to be decomposed are also symmetric. For the SVD of a symmetric matrix, using the target words $\bm{U}_k\bm{\Sigma}_k^{1/2}$ for word embeddings is equivalent to using the context words $\bm{V}_k\bm{\Sigma}_k^{1/2}$ for word embeddings. We vary the number of dimensions $k$ from 2, 50, 100, 200, $\cdots$, 1,000, 2,000, $\cdots$, 10,000.

\subsection{GloVe and SGNS}

We use the public implementation by \citet{pennington2014glove} to perform GloVe and choose the default hyperparameters. \citet{pennington2014glove} proposed to use the context vectors $\bm{o}_j$ in addition to target word vectors $\bm{e}_i$. Here, we only use target word vectors $\bm{e}_i$, set window size to 2 and set vocab minimum count to 100 for Text8, 500 for BNC, and 10,000 for Wiki052024, in the same way as for the SVD-based methods to keep the settings consistent. We vary the dimension $k$ of word embeddings from 200 to 600 with intervals of 100.

We use the public implementation by \citet{mikolov2013distributed} to perform SGNS, and use the vocabulary created by GloVe as the input of SGNS. We choose the default values except for the dimensions $k$ of word embeddings and  window size, which are chosen in the same way as in GloVe, to keep the settings consistent.

\subsection{BERT}\label{sub: setupbert}
We use two BERT models in our experiments: pre-trained BERT$_{\text{BASE}}$ and fine-tuned BERT$_{\text{BASE}}$. We extract the word embeddings from the last layer as well as the first layer, because \citet{bommasani2020interpreting} show that there is a clear preference towards the first quarter of the model layers. To be consistent with SVD-based methods, we use the same vocabulary constructed from Wiki052024 as in SVD-based methods.

Fine-tuned BERT$_{\text{BASE}}$ is the fine-tuning of BERT$_{\text{BASE}}$ on Wiki052024. Specifically, because BERT$_{\text{BASE}}$ is a very large neural network model containing about 110 million parameters and computationally it will be very expensive to run on a large corpus, thus we choose a sample from the Wiki052024 to fine-tune BERT$_{\text{BASE}}$. That is, we choose 0.02\% from the total 144,581,193 sentences. I.e., we use 28,916 sentences from Wiki052024. It is worth noting that in the fine-tuned BERT$_{\text{BASE}}$, we choose the max length for sentences is 128 instead of the default 512, because each sentence in Wiki052024 is an average of around 113 words. And we only use the MLM task instead of MLM and NSP, because 28,916 sentences are randomly chosen and are not necessary connecting the previous and next sentences. After we obtain the fine-tuned BERT$_{\text{BASE}}$, we extract word embedding from the last layer as well as first layer, and use the same vocabulary as in SVD-based methods.

\section{Results}\label{S: preres}

We make a distinction between conditions where no dimensionality reduction takes place, and conditions where dimensionality reduction is used. For no dimensionality reduction we compare TTEST, PMI, PPMI, WPMI, ROOT-TTEST, ROOTROOT-TTEST, STRATOS-TTEST. For dimensionality reduction we first compare CA with the more standard methods PMI-SVD, PPMI-SVD, PMI-GSVD, GloVe, SGNS, and then compare variants of CA as well as BERT.

\subsection{TTEST, PMI, PPMI, WPMI, ROOT-TTEST, ROOTROOT-TTEST, and STRATOS-TTEST}\label{Sub: fitresults}

First, we compare methods where no dimensionality reduction takes place. We show the Spearman’s correlation coefficient $\rho$ for the TTEST, PMI, PPMI, WPMI, ROOT-TTEST, ROOTROOT-TTEST, and STRATOS-TTEST matrices in Table~\ref{T: fittingmatrix} for Text8, BNC, and Wiki052024. The results for Text8 show that either ROOT-TTEST or ROOTROOT-TTEST is best across the four word similarity datasets. For BNC, ROOT-TTEST achieves the best performance across all four word similarity datasets. For Wiki052024, either ROOT-TTEST or PPMI performs best across the four word similarity datasets. In the Total block at the bottom of the table, we provide the sum of $\rho$-values for all four datasets. In the column of Total of Table~\ref{T: fittingmatrix}, we provide the sum of Spearman’s correlation coefficients over Text8, BNC, and Wiki052024. Overall, ROOT-TTEST perform best, followed by PPMI, and ROOTROOT-TTEST.

\begin{table}\footnotesize
\centering  
\caption{Text8, BNC, and Wiki052024: correlation coefficient $\rho$ for TTEST, PMI, PPMI, WPMI, ROOT-TTEST, ROOTROOT-TTEST, and STRATOS-TTEST matrices.} 
\label{T: fittingmatrix}
\begin{tabular}{lllllll}
\hline
&  & Text8 &BNC & Wiki052024 & Total
\\\hline 
\multirow{6}{*}{WordSim353}&TTEST & 0.588 & 0.427 & 0.515 & 1.531 \\
&PMI & 0.587 & 0.292 & 0.224 & 1.103 \\
&PPMI & 0.609 & 0.505 & \textbf{0.606} & 1.721 \\
&WPMI & 0.233 & 0.221 & 0.369 & 0.823 \\
&ROOT-TTEST & \textbf{0.658} & \textbf{0.539} & 0.588 & \textbf{1.785} \\
&ROOTROOT-TTEST & 0.646 & 0.495 & 0.507 & 1.648 \\
&STRATOS-TTEST & 0.438 & 0.314 & 0.438 & 1.190 \\
\hline 
\multirow{6}{*}{MEN}&TTEST & 0.248 & 0.260 & 0.208 & 0.717 \\
&PMI & 0.269 & 0.224 & 0.137 & 0.631 \\
&PPMI & 0.253 & 0.284 & \textbf{0.262} & 0.799 \\
&WPMI & 0.132 & 0.171 & 0.178 & 0.481 \\
&ROOT-TTEST & 0.305 & \textbf{0.293} & 0.251 & \textbf{0.850} \\
&ROOTROOT-TTEST & \textbf{0.317} & 0.263 & 0.236 & 0.816 \\
&STRATOS-TTEST & 0.156 & 0.130 & 0.141 & 0.427 \\
\hline 
\multirow{6}{*}{Turk}&TTEST & 0.619 & 0.649 & 0.567 & 1.835 \\
&PMI & 0.629 & 0.514 & 0.414 & 1.557 \\
&PPMI & 0.651 & 0.625 & 0.636 & 1.912 \\
&WPMI & 0.343 & 0.417 & 0.477 & 1.238 \\
&ROOT-TTEST & 0.666 & \textbf{0.659} & \textbf{0.657} & \textbf{1.982} \\
&ROOTROOT-TTEST & \textbf{0.667} & 0.616 & 0.620 & 1.903 \\
&STRATOS-TTEST & 0.561 & 0.525 & 0.551 & 1.637 \\
\hline 
\multirow{6}{*}{SimLex-999}&TTEST & 0.220 & 0.230 & 0.251 & 0.701 \\
&PMI & 0.257 & 0.168 & 0.184 & 0.609 \\
&PPMI & 0.251 & 0.277 & \textbf{0.352} & 0.880 \\
&WPMI & 0.139 & 0.118 & 0.192 & 0.449 \\
&ROOT-TTEST & \textbf{0.276} & \textbf{0.280} & 0.343 & \textbf{0.898} \\
&ROOTROOT-TTEST & 0.271 & 0.239 & 0.304 & 0.813 \\
&STRATOS-TTEST & 0.181 & 0.125 & 0.263 & 0.569 \\
\hline   
\multirow{6}{*}{Total}&TTEST & 1.675 & 1.566 & 1.542 & 4.783 \\
&PMI & 1.743 & 1.198 & 0.959 & 3.900 \\
&PPMI & 1.764 & 1.691 & \textbf{1.857} & 5.312 \\
&WPMI & 0.847 & 0.927 & 1.217 & 2.991 \\
&ROOT-TTEST & \textbf{1.904} & \textbf{1.772} & 1.839 & \textbf{5.515} \\
&ROOTROOT-TTEST & 1.901 & 1.613 & 1.666 & 5.179 \\
&STRATOS-TTEST & 1.336 & 1.093 & 1.393 & 3.822 \\
\hline 
\end{tabular}  
\end{table} 

\subsection{CA, PMI-SVD, PPMI-SVD, PMI-GSVD, GloVe, and SGNS}\label{sub: preliminary results}

Next, we compare CA (RAW-CA in Table~\ref{T: wordembeddings}) with the PMI-based methods PMI-SVD, PPMI-SVD, PMI-GSVD, GloVe, and SGNS for Text8, BNC, and Wiki052024. Table~\ref{T: wordembeddings} has a left part, where $p = 0$, and a right part, where $p = 0.5$. As $p$ does not exist in GloVe and SGNS, these methods have identical values for $p = 0$ and $p = 0.5$. Plots for $\rho$ as a function of $k$ for SVD-based methods about Text8, BNC, and Wiki052024 are in Supplementary materials B.

Comparing the last block of Table~\ref{T: wordembeddings} with the last block of Table~\ref{T: fittingmatrix} reveals that, overall, dimensionality reduction is beneficial for the size of $\rho$, as CA, PMI-SVD, PPMI-SVD, PMI-GSVD, ROOT-CA, ROOTROOT-CA, and ROOT-CCA do better than their respective counterparts TTEST, PMI, PPMI, WPMI, ROOT-TTEST, ROOTROOT-TTEST, and STRATOS-TTEST.

For an overall comparison of the dimensionality reduction methods, we study the block at the bottom of Table~\ref{T: wordembeddings}, which provides the sum of the $\rho$-values over the four word similarity datasets. In the column of Total of Table~\ref{T: wordembeddings}, we provide the sum of Spearman’s correlation coefficients over Text8, BNC, and Wiki052024. For both $p = 0$ and $p = 0.5$, among RAW-CA, PMI-SVD, PPMI-SVD, PMI-GSVD, GloVe, and SGNS, overall PPMI-SVD performs best, closely followed by PMI-SVD and SGNS. RAW-CA, PMI-GSVD, and GloVe follow at some distance.
The popular method GloVe does not perform well especially in Text8 and BNC. Possibly the conditions of the study are not optimal for GloVe, as the Text8 and BNC corpora are, with 11,815 and 11,332 terms respectively, possibly too small to obtain reliable results \citep{jiang-etal-2018-learning}.

\begin{table}
\caption{Text8, BNC, and Wiki052024: correlation coefficient $\rho$ for SVD-based methods with $p = 0, 0.5$ and for GloVe and SGNS.} 
\label{T: wordembeddings}
\centering
\setlength{\tabcolsep}{3pt}
{\fontsize{6.8}{8}\selectfont
\begin{tabular}{ll|lllllll|lllllll}
\hline
   & &\multicolumn{6}{c}{$p = 0$} & & \multicolumn{6}{c}{$p = 0.5$} 
\\   & &\multicolumn{2}{c}{Text8} &\multicolumn{2}{c}{BNC}&\multicolumn{2}{c}{Wiki052024} & &\multicolumn{2}{c}{Text8} &\multicolumn{2}{c}{BNC} &\multicolumn{2}{c}{Wiki052024}
\\
&  &$k$&$\rho$&$k$&$\rho$&$k$&$\rho$&Total&$k$&$\rho$&$k$&$\rho$&$k$&$\rho$&Total\\\hline 
\multirow{8}{*}{Wor353} &RAW-CA & 600 & 0.578 & 400 & 0.465 & 800 & 0.493 & 1.536 & 9000 & 0.609 & 10000 & 0.498 & 6000 & 0.535 & 1.643 \\
&PMI-SVD & 400 & 0.675 & 600 & 0.628 & 200 & 0.624 & 1.927 & 400 & 0.683 & 500 & 0.579 & 500 & 0.597 & 1.859 \\
&PPMI-SVD & 400 & 0.681 & 700 & 0.628 & 700 & 0.672 & 1.981 & 200 & 0.694 & 2000 & 0.623 & 900 & \textbf{0.688} & 2.005 \\
&GloVe & 200 & 0.422 & 600 & 0.522 & 600 & 0.663 & 1.607 & 200 & 0.422 & 600 & 0.522 & 600 & 0.663 & 1.607 \\
&SGNS & 300 & 0.668 & 600 & 0.551 & 600 & 0.654 & 1.873 & 300 & 0.668 & 600 & 0.551 & 600 & 0.654 & 1.873 \\
&PMI-GSVD & 700 & 0.512 & 600 & 0.468 & 400 & 0.572 & 1.552 & 6000 & 0.548 & 3000 & 0.449 & 900 & 0.586 & 1.583 \\
&ROOT-CA & 300 & 0.668 & 400 & 0.623 & 500 & 0.669 & 1.960 & 500 & 0.688 & 900 & \textbf{0.657} & 600 & 0.682 & \textbf{2.027} \\
&ROOTROOT-CA & 200 & \textbf{0.692} & 200 & \textbf{0.635} & 800 & \textbf{0.700} & \textbf{2.027} & 300 & \textbf{0.697} & 400 & 0.630 & 800 & 0.679 & 2.005 \\
&ROOT-CCA & 100 & 0.682 & 700 & 0.627 & 400 & 0.667 & 1.977 & 300 & 0.684 & 600 & 0.620 & 600 & 0.681 & 1.984 \\
\hline
\multirow{8}{*}{MEN}&RAW-CA & 300 & 0.223 & 600 & 0.293 & 1000 & 0.190 & 0.706 & 7000 & 0.256 & 9000 & 0.299 & 9000 & 0.242 & 0.798 \\
&PMI-SVD & 800 & 0.328 & 700 & 0.393 & 1000 & 0.315 & 1.036 & 600 & 0.317 & 2000 & 0.357 & 6000 & 0.288 & 0.962 \\
&PPMI-SVD & 800 & 0.336 & 500 & 0.394 & 900 & 0.357 & 1.087 & 800 & 0.324 & 1000 & 0.358 & 9000 & \textbf{0.327} & 1.009 \\
&GloVe & 300 & 0.175 & 600 & 0.310 & 500 & 0.326 & 0.811 & 300 & 0.175 & 600 & 0.310 & 500 & 0.326 & 0.811 \\
&SGNS & 400 & 0.295 & 400 & 0.333 & 600 & 0.318 & 0.945 & 400 & 0.295 & 400 & 0.333 & 600 & 0.318 & 0.945 \\
&PMI-GSVD & 800 & 0.267 & 600 & 0.318 & 400 & 0.269 & 0.854 & 5000 & 0.256 & 3000 & 0.308 & 3000 & 0.270 & 0.834 \\
&ROOT-CA & 800 & 0.325 & 500 & \textbf{0.400} & 400 & 0.311 & 1.036 & 9000 & 0.324 & 800 & \textbf{0.374} & 8000 & 0.313 & \textbf{1.012} \\
&ROOTROOT-CA & 600 & \textbf{0.340} & 400 & 0.396 & 900 & \textbf{0.360} & \textbf{1.096} & 1000 & \textbf{0.332} & 4000 & 0.359 & 10000 & 0.318 & 1.009 \\
&ROOT-CCA & 600 & 0.315 & 400 & 0.392 & 700 & 0.313 & 1.019 & 900 & 0.298 & 800 & 0.355 & 8000 & 0.287 & 0.940 \\
\hline
\multirow{8}{*}{Turk}&RAW-CA & 400 & 0.549 & 100 & 0.562 & 600 & 0.497 & 1.608 & 400 & 0.592 & 10000 & 0.588 & 7000 & 0.537 & 1.718 \\
&PMI-SVD & 100 & 0.656 & 50 & 0.652 & 50 & 0.595 & 1.903 & 300 & \textbf{0.677} & 500 & 0.661 & 700 & 0.597 & 1.935 \\
&PPMI-SVD & 50 & 0.668 & 50 & 0.671 & 50 & 0.641 & 1.980 & 50 & \textbf{0.677} & 50 & 0.683 & 800 & 0.666 & \textbf{2.027} \\
&GloVe & 600 & 0.502 & 200 & 0.540 & 200 & \textbf{0.692} & 1.733 & 600 & 0.502 & 200 & 0.540 & 200 & \textbf{0.692} & 1.733 \\
&SGNS & 200 & 0.651 & 300 & 0.650 & 200 & 0.659 & 1.961 & 200 & 0.651 & 300 & 0.650 & 200 & 0.659 & 1.961 \\
&PMI-GSVD & 900 & 0.495 & 200 & 0.506 & 100 & 0.501 & 1.501 & 5000 & 0.563 & 10000 & 0.584 & 6000 & 0.587 & 1.734 \\
&ROOT-CA & 50 & 0.649 & 50 & \textbf{0.695} & 100 & 0.662 & \textbf{2.006} & 100 & 0.661 & 50 & \textbf{0.684} & 700 & 0.680 & 2.025 \\
&ROOTROOT-CA & 50 & \textbf{0.669} & 50 & 0.666 & 100 & 0.666 & 2.001 & 50 & 0.664 & 300 & 0.673 & 1000 & 0.668 & 2.005 \\
&ROOT-CCA & 50 & 0.633 & 50 & 0.672 & 50 & 0.667 & 1.972 & 100 & 0.665 & 100 & 0.678 & 100 & 0.670 & 2.013 \\
\hline
\multirow{8}{*}{Sim999}&RAW-CA & 4000 & 0.219 & 2000 & 0.322 & 10000 & 0.289 & 0.830 & 8000 & 0.243 & 7000 & 0.327 & 10000 & 0.315 & 0.886 \\
&PMI-SVD & 700 & 0.310 & 900 & 0.409 & 900 & 0.411 & 1.130 & 3000 & 0.315 & 900 & 0.372 & 5000 & 0.414 & 1.101 \\
&PPMI-SVD & 700 & 0.309 & 500 & 0.393 & 2000 & \textbf{0.463} & 1.165 & 3000 & 0.308 & 500 & 0.368 & 4000 & \textbf{0.438} & 1.114 \\
&GloVe & 500 & 0.148 & 500 & 0.255 & 600 & 0.384 & 0.787 & 500 & 0.148 & 500 & 0.255 & 600 & 0.384 & 0.787 \\
&SGNS & 600 & 0.306 & 400 & 0.376 & 600 & 0.404 & 1.086 & 600 & 0.306 & 400 & 0.376 & 600 & 0.404 & 1.086 \\
&PMI-GSVD & 900 & 0.272 & 4000 & 0.365 & 4000 & 0.356 & 0.993 & 5000 & 0.271 & 3000 & 0.312 & 5000 & 0.370 & 0.953 \\
&ROOT-CA & 2000 & 0.295 & 900 & 0.415 & 3000 & 0.430 & 1.140 & 5000 & 0.309 & 2000 & \textbf{0.395} & 8000 & 0.433 & \textbf{1.137} \\
&ROOTROOT-CA & 700 & \textbf{0.321} & 900 & 0.410 & 3000 & 0.445 & \textbf{1.176} & 700 & \textbf{0.317} & 900 & 0.376 & 10000 & 0.417 & 1.110 \\
&ROOT-CCA & 1000 & 0.294 & 1000 & \textbf{0.421} & 2000 & 0.439 & 1.154 & 7000 & 0.303 & 2000 & 0.391 & 8000 & 0.428 & 1.121 \\
\hline
\multirow{8}{*}{Total}
&RAW-CA &  & 1.569 &  & 1.642 &  & 1.469 & 4.680 &  & 1.701 &  & 1.713 &  & 1.629 & 5.044 \\
&PMI-SVD &  & 1.969 &  & 2.082 &  & 1.945 & 5.996 &  & 1.993 &  & 1.968 &  & 1.895 & 5.857 \\
&PPMI-SVD &  & 1.994 &  & 2.087 &  & 2.133 & 6.213 &  & 2.003 &  & 2.032 &  & \textbf{2.119} & 6.154 \\
&GloVe &  & 1.246 &  & 1.627 &  & 2.065 & 4.938 &  & 1.246 &  & 1.627 &  & 2.065 & 4.938 \\
&SGNS &  & 1.920 &  & 1.911 &  & 2.035 & 5.865 &  & 1.920 &  & 1.911 &  & 2.035 & 5.865 \\
&PMI-GSVD &  & 1.546 &  & 1.657 &  & 1.698 & 4.901 &  & 1.638 &  & 1.653 &  & 1.813 & 5.104 \\
&ROOT-CA &  & 1.936 &  & \textbf{2.133} &  & 2.073 & 6.142 &  & 1.982 &  & \textbf{2.110} &  & 2.108 & \textbf{6.201} \\
&ROOTROOT-CA &  & \textbf{2.022} &  & 2.106 &  & \textbf{2.172} & \textbf{6.300} &  & \textbf{2.009} &  & 2.037 &  & 2.082 & 6.129 \\
&ROOT-CCA &  & 1.924 &  & 2.112 &  & 2.085 & 6.122 &  & 1.949 &  & 2.044 &  & 2.066 & 6.059 \\
\hline
\end{tabular}%
}
\end{table}

\begin{table}
    \centering
        \caption{Text8: the number of extreme values}
    \label{T: text8numext}
    \begin{tabular}{lrrrrrr}
    \hline
        &  LT$f_1$& GT$f_3$ & total \\
    \hline
PMI           &     4,335 &   27,984 &    32,319  \\
PPMI          &        0 &   27,984 &    27,984  \\
WPMI          &  1,038,236 &  345,995 &  1,384,231 \\
\hline
TTEST      &    50,560 &  627,046 &   677,606  \\
ROOT-TTEST     &     5,985 &  448,860 &   454,845  \\
ROOTROOT-TTEST &     4,942 &  396,437 &   401,379  \\
STRATOS-TTEST  &        0 &  400,703 &   400,703 \\
    \hline
    \end{tabular}
\end{table}

\begin{figure}
    \centering
    \includegraphics[width=1\textwidth]{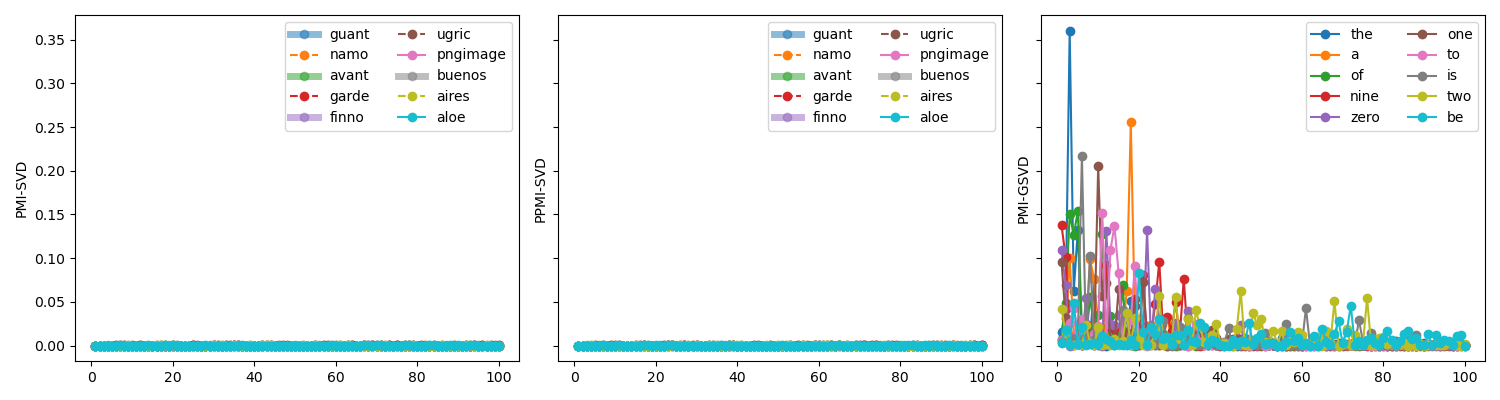}
    \vspace{3mm}
         \caption{Text8: the contribution of the rows, corresponding to the top 10 extreme values, to the first 100 dimensions of PMI-SVD, PPMI-SVD, PMI-GSVD.}
    \label{F: text8pmiextremevaluecontri}
\end{figure}

\begin{table}
\centering
{
\fontsize{8}{8}\selectfont
\caption{Wiki052024: correlation coefficient $\rho$ for BERT.} 
\label{T: wordembeddingswiki}
\begin{tabular}{ll|ll}
\hline
& & $\rho$ 
\\\hline
\multirow{4}{*}{WordSim353}
&BERT-PRETRAIN-LSAT-LAYER &0.303\\
&BERT-PRETRAIN-FIRST-LAYER &\textbf{0.700} \\
&BERT-FINETUNE-LSAT-LAYER &0.373 \\
&BERT-FINETUNE-FIRST-LAYER&0.695 \\
\hline
\multirow{4}{*}{MEN}
&BERT-PRETRAIN-LSAT-LAYER &0.176 \\
&BERT-PRETRAIN-FIRST-LAYER  & 0.364  \\
&BERT-FINETUNE-LSAT-LAYER&0.210 \\
&BERT-FINETUNE-FIRST-LAYER&\textbf{0.365}\\
\hline
\multirow{4}{*}{Turk}
&BERT-PRETRAIN-LSAT-LAYER&0.280 \\
&BERT-PRETRAIN-FIRST-LAYER  & \textbf{0.642} \\
&BERT-FINETUNE-LSAT-LAYER &0.348 \\
&BERT-FINETUNE-FIRST-LAYER &0.623 \\

\hline
\multirow{4}{*}{SimLex-999}
&BERT-PRETRAIN-LSAT-LAYER &0.174 \\
&BERT-PRETRAIN-FIRST-LAYER  & 0.486 \\
&BERT-FINETUNE-LSAT-LAYER &0.240 \\
&BERT-FINETUNE-FIRST-LAYER &\textbf{0.492} \\
\hline
\multirow{4}{*}{Total}
&BERT-PRETRAIN-LSAT-LAYER & 0.933\\
&BERT-PRETRAIN-FIRST-LAYER& \textbf{2.191}\\
&BERT-FINETUNE-LSAT-LAYER&1.171\\
&BERT-FINETUNE-FIRST-LAYER  &2.176\\
\hline
\end{tabular}}
\end{table}

As the focus in this paper is on the performance of CA, we give some extra attention to RAW-CA and the similar PMI-GSVD. Even though CA and PMI-GSVD have the same weighting function $p_{i+}p_{+j}$, and should be close when $p_{ij}/\left(p_{i+}p_{+j}\right) - 1$ is small (compare the discussion around Equations~(\ref{approxcapmi}, \ref{objectiveCAalter})) their performances are rather different. This may be because there are extremely large values (larger than 35,000) in the fitting function $\left(p_{ij}/\left(p_{i+}p_{+j}\right)-1\right)$ of CA, which makes the fitting function of CA not close to the fitting function log $\left(p_{ij}/\left(p_{i+}p_{+j}\right)\right)$ of PMI-GSVD.

When we compare PMI-GSVD with PMI-SVD, we are surprised to find that weighting rows and columns appears to decrease the values of $\rho$. This is in contrast with the reliability principle of \citet{salle2022understanding} discussed above.

We now discuss why PMI-SVD and PPMI-SVD do better than PMI-GSVD. It turns out that the number and sizes of extreme values in the matrix WPMI decomposed by PMI-GSVD are much larger than in PMI and PPMI, and this results in PMI-GSVD dimensions being dominated by single words. We only include non-zero elements in the PMI matrix as the PMI matrix is sparse: 94.2\% of the entries are
zero for Text8; for a fair comparison, the corresponding 94.2\% of entries in the PPMI and WPMI matrices are also ignored. Following box plot methodology \citep{tukey1977exploratory, schwertman2004simple, dodge2008concise}, extreme values are determined as follows: let $q_1$ and $q_3$ be the first and third sample quartiles, and let $f_1 = q_1 - 1.5(q_3 - q_1)$, $f_3 = q_3 + 1.5 (q_3 - q_1)$. Then extreme values are defined as values less than $f_1$ (LT$f_1$) or greater than $f_3$ (GT$f_3$). The first three rows in  Table~\ref{T: text8numext} show the number of extreme elements in the PMI, PPMI, WPMI matrices. The number of extreme values of the WPMI matrix (1,384,231) is much larger than that of PMI and PPMI (32,319 and 27,984). 
Furthermore, in WPMI the extremeness of values is much larger than in PMI and PPMI. Let the averaged contribution of each cell, expressed as a proportion, be $1/\left(11,815\times 11,815\right)$. However, in WPMI, the most extreme entry, found for (the, the),
contributes around 0.01126 to the total inertia.
In PMI (PPMI) the most extreme entry is (guant, namo) or (namo, guant) and contributes around $3.1\times 10^{-6}$ ($3.2\times 10^{-6}$) to the total inertia. Figure~\ref{F: text8pmiextremevaluecontri} shows the contribution of the rows for the corresponding to top 10 extreme values, to the first 100 dimensions of PMI-SVD, PPMI-SVD, PMI-GSVD. The rows, corresponding to the top extreme values in the WPMI matrix, take up a much bigger contribution to the first dimensions of PMI-GSVD. For example, in PMI-GSVD, the “the” row contributes more than 0.3 to the third dimension, while in PMI-SVD and PPMI-SVD, the contributions are much more even. Thus the PMI-GSVD solution is hampered by extreme cells in the WPMI matrix that is decomposed. Similar results can be found for BNC and Wiki052024 in Supplementary materials C and D, respectively.

\subsection{The results for three variants of CA as well as BERT}\label{S: resrootrootca}

Now we compare the three variants of CA (ROOT-CA, ROOTROOT-CA, ROOT-CCA) with CA-RAW and the winner of the PMI-based methods, PPMI-SVD.

First, in Table \ref{T: wordembeddings} for Text8, BNC, and Wiki052024, the three variants of CA perform much better than RAW-CA in each word similarity dataset and each corpus, both for $p = 0$ and $p = 0.5$. In the block at the bottom of Table \ref{T: wordembeddings}, overall, the performance of the three variants is similar, where ROOT-CA and ROOTROOT-CA outperform ROOT-CCA slightly.

\begin{figure}
    \centering
    \includegraphics[width=1\textwidth]{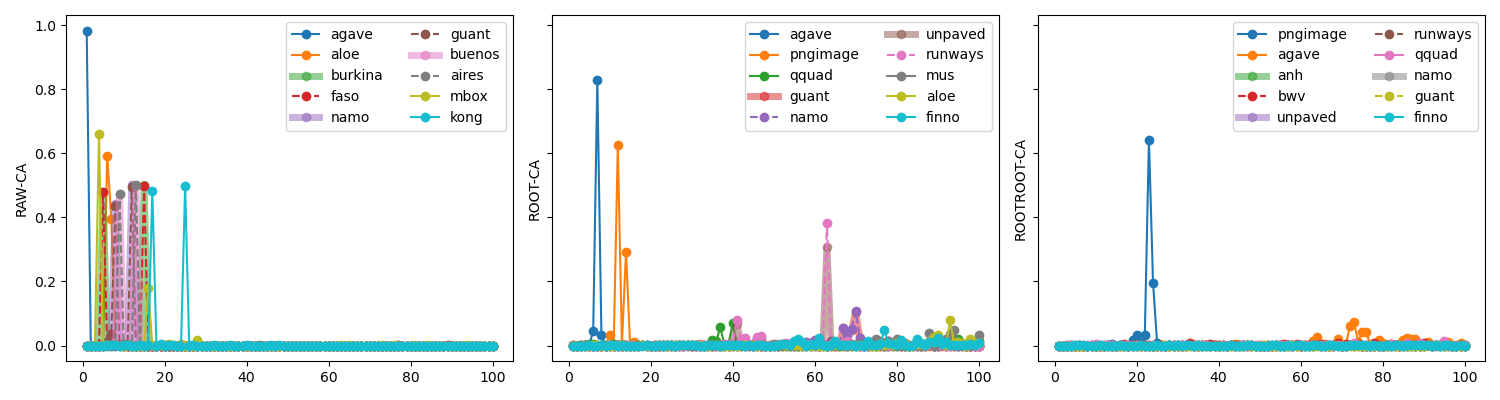}
             \vspace{3mm}
         \caption{Text8: the contribution of the rows, corresponding to top 10 extreme values, to first 100 dimensions of RAW-CA, ROOT-CA, ROOTROOT-CA.}
    \label{F: text8caextremevaluecontri}
\end{figure}

The lower part of Table~\ref{T: text8numext} shows the number of extreme values of TTEST, ROOT-TTEST, ROOTROOT-TTEST and STRATOS-TTEST matrices for Text8.  Similar with PMI, PPMI, WPMI, 94.2\% of entries are ignored. The number of extreme values of the TTEST matrix (677,606) is  larger than that of ROOT-TTEST, ROOTROOT-TTEST and STRATOS-TTEST (454,845, 401,379, and 400,703). Furthermore, in TTEST the extremeness of the extreme values is larger than those in ROOT-TTEST, ROOTROOT-TTEST, and STRATOS-TTEST. For example, in TTEST the most extreme entry (agave, agave) contributes around 0.02117 to the total inertia, while in ROOT-TTEST, ROOTROOT-TTEST, and STRATOS-TTEST, the most extreme entries (agave, agave), (pngimage, pngimage), and (agave, agave) contribute around 0.00325, 0.00119, and 0.00017, respectively. Figure~\ref{F: text8caextremevaluecontri} shows the contribution of the rows for the top 10 extreme values, to the first 100 dimensions of RAW-CA, ROOT-CA, and ROOTROOT-CA (The corresponding plot about ROOT-CCA is in Supplementary materials E). In RAW-CA, the rows, corresponding to top extreme values of
TTEST, take up a big contribution to the first dimensions of RAW-CA. For example, in RAW-CA, the “agave” row contributes around 0.983 to the first dimension, while in ROOT-CA and ROOTROOT-CA, the contributions are much smaller which also holds for ROOT-CCA. Similar results can be found for BNC and Wiki052024 in Supplementary materials F and G, respectively. Thus, we infer that the extreme values in TTEST are the important reason that RAW-CA performs badly.

Second, in the rows of the block at the bottom of Table~\ref{T: wordembeddings}, the overall performances of ROOT-CA, ROOTROOT-CA, ROOT-CCA are comparable to or sometimes slightly better than PPMI-SVD. Specifically, for both $p = 0$ and $p = 0.5$, ROOTROOT-CA and ROOT-CA achieve the highest $\rho$ for Text8 and BNC
corpora, respectively. For Wiki052024, when $p = 0$, ROOTROOT-CA performs best with $\rho = 2.172$; when 
$p = 0.5$, PPMI-SVD achieves the highest value with
$\rho = 2.119$. Notably, 2.172 is larger than 2.119, indicating ROOTROOT-CA at $p = 0$ yields the best overall performance. Based on these results, no matter what we know about the corpus, ROOTROOT-CA and ROOT-CA appear to have potential to improve the performance in NLP tasks.

Comparied with BERT shown in Table~\ref{T: wordembeddingswiki}, BERT$_{\text{BASE}}$ in the last layer is worst,  but in the first layer is best. Although BERT in the first layer is the best, ROOTROOT-CA and ROOT-CA provide a competitive result, such as, ROOTROOT-CA with score 2.735 VS BERT$_{\text{BASE}}$ 2.827. In addition, ROOT-CA and ROOTROOT-CA are better than BERT in the Turk dataset, which indicates that different methods are suitable to different datasets.

\section{Discussion}\label{S: discussion}

PMI is an important concept in natural language processing. In this paper, we theoretically compare CA with three PMI-based methods with respect to their objective functions. CA is a weighted factorization of a matrix where the fitting function is $\left(p_{ij}/\left(p_{i+}p_{+j}\right)-1\right)$ and the weighting function is the product of row margins and column margins $p_{i+}p_{+j}$. When the elements in the fitting function $\left(p_{ij}/\left(p_{i+}p_{+j}\right)-1\right)$ of CA are small, CA is close to a weighted factorization of the PMI matrix where the weighting function is the product $p_{i+}p_{+j}$. This is because $\left(p_{ij}/\left(p_{i+}p_{+j}\right)-1\right)$ is close to $\text{log}\left(p_{ij}/\left(p_{i+}p_{+j}\right)\right)$ when $\left(p_{ij}/\left(p_{i+}p_{+j}\right)-1\right)$ is small. This theoretical link places CA and PMI-based approaches within a unified analytical framework.

Extracted word-context matrices are prone to overdispersion. To remedy the overdispersion, we presented ROOTROOT-CA. That is, we perform CA on the root-root transformation of the word-context matrix. We also apply CA to the square-root transformation of the word-context matrix (ROOT-CA) to stabilize the variance of the matrix entries. In addition, we present ROOT-CCA, described in \citet{stratos2015model}, which is similar with ROOT-CA. The empirical comparison on word similarity tasks shows that ROOTROOT-CA achieves the best overall results in the Text8 corpus and Wiki052024 corpus, and ROOT-CA achieves the best overall results in the BNC corpus. Overall, the performance of ROOT-CA and ROOTROOT-CA is slightly better than the performance of PMI-based methods. In CA-based methods, it is no logarithm, which avoid the illness for 0 element compared with PMI. Therefore, CA-based methods could become an alternative of PMI-based methods, and facilitate the development of NLP.

A key explanation for the performance differences lies in the influence of extreme values. PMI-GSVD performs worse than PMI-SVD/PPMI-SVD and CA performs worse than ROOT-CA, ROOTROOT-CA, and ROOT-CCA, because the extreme values take up large contribution to the first dimensions of PMI-GSVD and RAW-CA. This gives a direction to improve the performance of SVD-based methods, such as by removing the effect of extreme values.

Interesting, our results show PMI-SVD performs better than PMI-GSVD, where PMI-SVD has equally weighting function in the objective function with 1 and PMI-GSVD has unequal weighting function with the product of the marginal frequencies of word and of context. This phenomenon is in contrast with the reliability principle, proposed by \cite{salle2022understanding}. The reliability principle refers to that the objective function should have a weight on the reconstruction error that is a monotonically increasing function of the marginal frequencies of word and of context. This phenomenon in our results inspires researcher to rethink about the theory of NLP.

Although transformer-based encoders such as BERT are dominant in modern NLP, our results show that traditional static word embedding methods remain competitive. The results on word similarity tasks for Wiki052024 corpus show the overall results of ROOT-CA and ROOTROOT-CA are competitive with that of BERT. ROOT-CA and ROOTROOT-CA are better than BERT in the Turk dataset, which indicates that different methods are suitable for different datasets. This also suggests that simpler methods can sometimes outperform more complex models for highly specialized tasks.

Moreover, static embedding methods continue to play an important role in NLP for several reasons. First, different models are suitable for different tasks and different datasets. For example, \cite{aida2021comprehensive} show PMI-based methods are better than BERT in measuring semantic differences across corpora.
Second, integrating static word embedding models and transformer-based models can improve the performance of both \citep{Enhancing2025Adam}. Third, the transformer-based models are often less interpretable \citep{lai-dang2025adaptive}. This lack of explainability is significantly limiting used in domains where decisions are critical such as the medical and legal fields. In contrast, the co-occurrence-based model is more explainable. Last, transformer-based models require substantial computational resources and large corpora \citep{10.1145/3773076}, making traditional embedding methods more suitable for low-resource settings.
Overall, static word embeddings are not outdated and are still active in NLP.

In this paper, we explore ROOT-CA and ROOTROOT-CA, where ROOT-CA uses a power of 0.5 of the original elements $x_{ij}$ while ROOTROOT-CA uses 0.25 of the original elements $x_{ij}$. Our aim was to study the performance of CA with respect to the other methods, and for this purpose, focusing on values 0.25 and 0.5 was sufficient. It may be of interest to study a general power transformation $x_{ij}^\delta$ (or other power versions such as $(p_{ij}/(p_{i+}p_{+j}))^\delta$) where $\delta$ could range between any two non-negative values \citep{CUADRAS200664, greenacre2009power, Beh2023, beh2024correspondence}. Here 0.5 and 0.25 are special cases of this general transformation. 

\section{Conclusion}\label{S: conclusion}

In this paper, we provided a unified theoretical and empirical comparison between correspondence analysis (CA) and PMI-based methods for static word embeddings, and introduced two CA variants, ROOT-CA and ROOTROOT-CA, to the NLP domain. We showed that CA can be interpreted as a weighted factorization of PMI matrix when deviations from independence are small, thereby clarifying the connection of CA-, PMI-, and SVD-based methods. To mitigate overdispersion in word–context matrices, we proposed ROOTROOT-CA based on power transformation of the matrix entries; to stabilize variance, we proposed ROOT-CA. Experiments on multiple corpora demonstrate that these variants consistently achieve competitive and often superior performance compared with PMI-based methods, while remaining competitive with transformer-based models on word similarity tasks. The analysis found that the performance of SVD-based methods (including CA, ROOT-CA, ROOTROOT-CA, PMI-SVD, PPMI-SVD, and PMI-GSVD) is largely influenced by extreme values in decomposed matrices before SVD. Overall, our results highlight and suggest that ROOT-CA and ROOTROOT-CA are effective alternatives for NLP applications, and inspire researchers to improve SVD-based methods by controlling extreme values of the decomposed matrix. The code is available at \url{https://github.com/qianqianqi28/ca-pmi-word-embeddings}
.

\section*{Data availability}

The Text8 corpus and BNC corpus that support the findings of this study are openly available by \citet{text8} and \citet{BNC}, respectively. Wiki052024 is from the "enwiki-20240501-pages-articles-multistream.xml.bz2" which is downloaded from \cite{Wiki052024}.

The four word similarity datasets for word similarity tasks are from \url{https://github.com/valentinp72/svd2vec/tree/master/svd2vec/datasets/similarities}.

\section*{Statements and Declarations}

\textbf{Competing Interests: }Author Qianqian Qi is supported by the China Scholarship Council. Author Ayoub Bagheri, Author David J. Hessen, and Author Peter G. M. van der Heijden declare none.

\section*{Acknowledgements}

Author Qianqian Qi is supported by the China Scholarship Council.

\section*{Author Contributions}
All authors contributed to the design and conception of the study. Author Qianqian Qi performed the experiments. All authors contributed to the final analysis and discussion, and reviewed and edited the text.

\bibliography{sn-bibliography.bib}

\appendixqq

\appendixqqsection{An alternative coordinates system for CA}\label{SS: altsys}

For row points $i$ and $i'$, with coordinates $\sigma_k\phi_{ik}$ and $\sigma_k\phi_{i'k}$ on dimension $k$ in $K-$dimensional space we have
\begin{equation}
\label{cosinesim}
\begin{split}
\text{cosine}(\text{row}_i, \text{row}_{i'})
& = \frac{\sum_{k=1}^{K}\left(\phi_{ik}\sigma_k\right)\left(\phi_{i'k}\sigma_k\right)}{\sqrt{\sum_{k=1}^{K}\left(\phi_{ik}\sigma_k\right)^2\cdot\sum_{k=1}^{K}\left(\phi_{i'k}\sigma_k\right)^2}}\\
& = \frac{\sum_{k=1}^{K}\left(p_{i+}^{-\frac{1}{2}}u_{ik}\sigma_k\right)\left(p_{i'+}^{-\frac{1}{2}}u_{i'k}\sigma_k\right)}{\sqrt{\sum_{k=1}^{K}\left(p_{i+}^{-\frac{1}{2}}u_{ik}\sigma_k\right)^2\cdot\sum_{k=1}^{K}\left(p_{i'+}^{-\frac{1}{2}}u_{i'k}\sigma_k\right)^2}}\\
& = \frac{\sum_{k=1}^{K}\left(u_{ik}\sigma_k\right)\left(u_{i'k}\sigma_k\right)}{\sqrt{\sum_{k=1}^{K}\left(u_{ik}\sigma_k\right)^2\cdot\sum_{k=1}^{K}\left(u_{i'k}\sigma_k\right)^2}},
\end{split}
\end{equation}
so the terms $p_{i+}^{-\frac{1}{2}}$ drop out of the equation. A similar result is found for column points. 

\appendixqqsection{Plots for $\rho$ as a function of $k$ for SVD-based methods}

Plots are for $\rho$ as a function of $k$ for SVD-based methods. 

\begin{figure}[H]
    \centering
        \begin{subfigure}[b]{0.42\linewidth}
         \centering
         \includegraphics[width=1\textwidth]{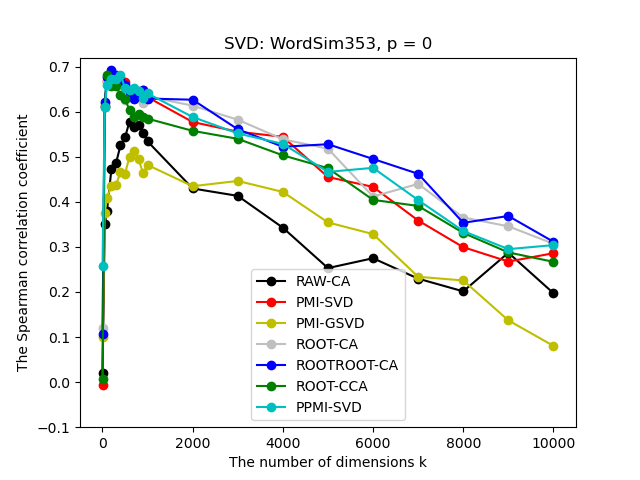}
\caption{WordSim353: $p = 0$}\label{F: text8SVDWordSim3530}
         \end{subfigure}
    \begin{subfigure}[b]{0.42\linewidth}
         \centering
         \includegraphics[width=1\textwidth]{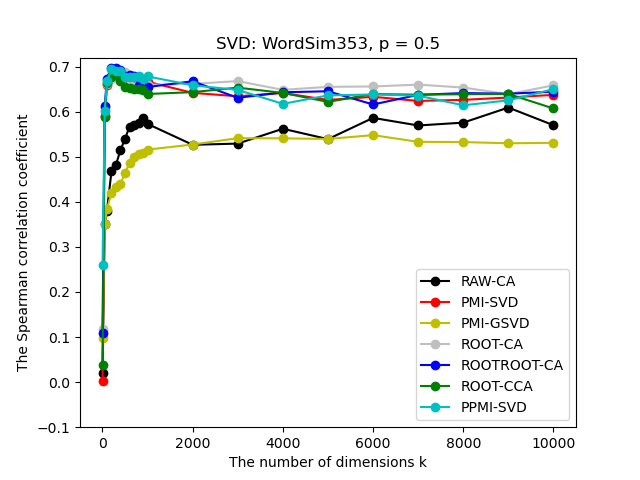}
\caption{WordSim353:$p = 0.5$}\label{F: text8SVDWordSim3530dot5}
         \end{subfigure}
    \begin{subfigure}[b]{0.42\linewidth}
         \centering
         \includegraphics[width=1\textwidth]{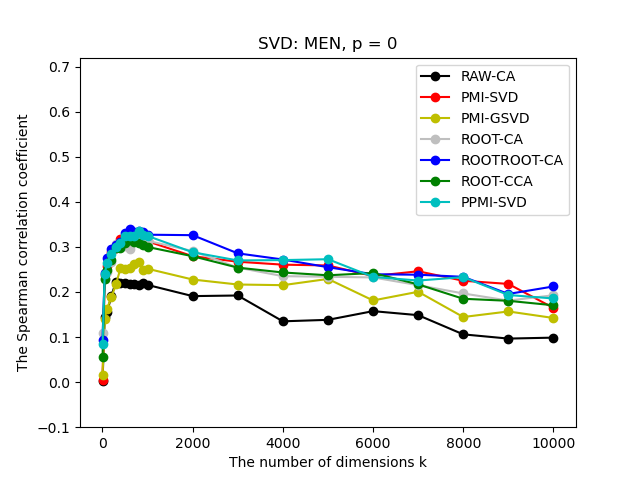}
\caption{MEN: $p = 0$}\label{F: text8SVDMEN0}
         \end{subfigure}
    \begin{subfigure}[b]{0.42\linewidth}
         \centering
         \includegraphics[width=1\textwidth]{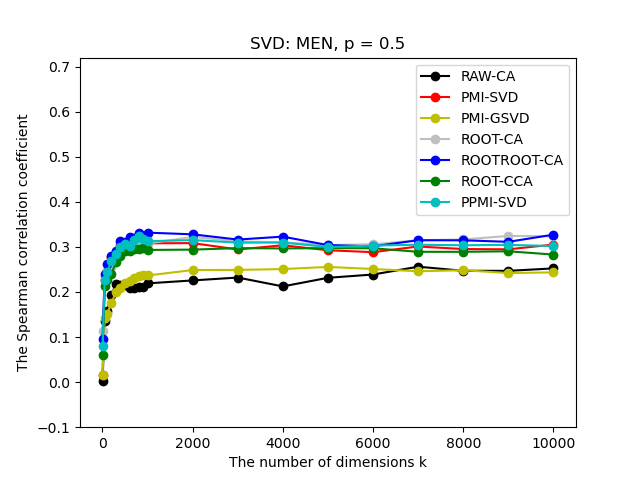}
\caption{MEN: $p = 0.5$}\label{F: text8SVDMEN0dot5}
         \end{subfigure}
                 \begin{subfigure}[b]{0.42\linewidth}
         \centering
         \includegraphics[width=1\textwidth]{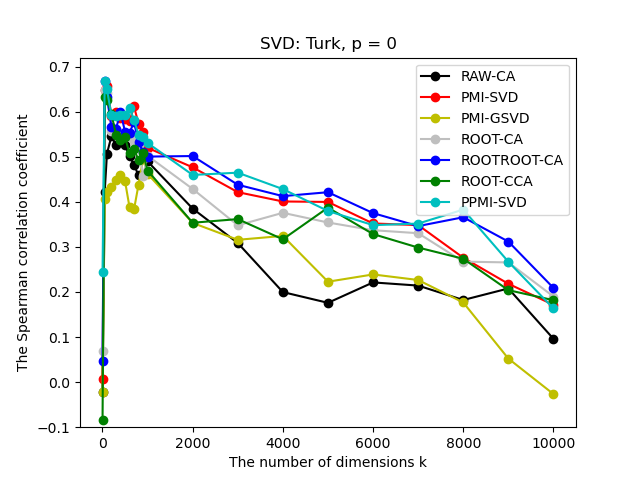}
\caption{Turk: $p = 0$}\label{F: text8SVDTurk0}
         \end{subfigure}
    \begin{subfigure}[b]{0.42\linewidth}
         \centering
         \includegraphics[width=1\textwidth]{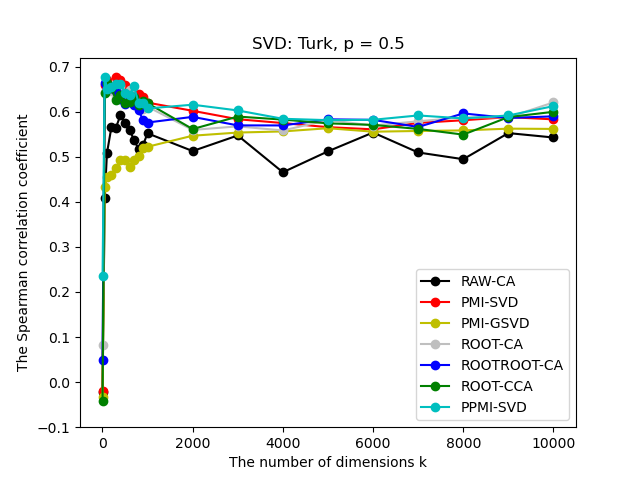}
\caption{Turk: $p = 0.5$}\label{F: text8SVDTurk0dot5}
         \end{subfigure}
        \caption{Text8}
    \label{F: text8SVD1}
\end{figure}
         
\begin{figure}[H]
 \ContinuedFloat
    \centering    
                 \begin{subfigure}[b]{0.42\linewidth}
         \centering
         \includegraphics[width=1\textwidth]{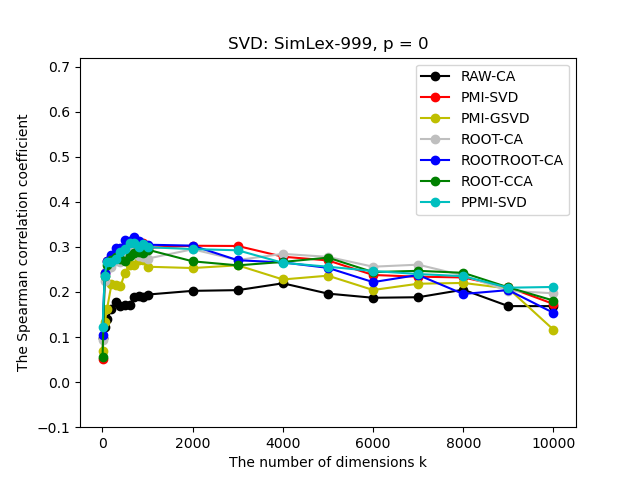}
\caption{SimLex-999: $p = 0$}\label{F: text8SVDSimLex-9990}
         \end{subfigure}
    \begin{subfigure}[b]{0.42\linewidth}
         \centering
         \includegraphics[width=1\textwidth]{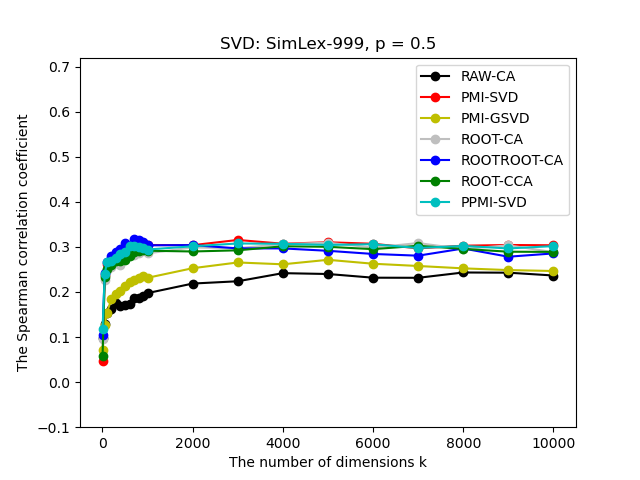}
\caption{SimLex-999: $p = 0.5$}\label{F: text8SVDSimLex-9990dot5}
         \end{subfigure}
         \caption{Text8}
    \label{F: text8SVD2}
\end{figure}

\begin{figure}[H]
    \centering
        \begin{subfigure}[b]{0.42\linewidth}
         \centering
         \includegraphics[width=1\textwidth]{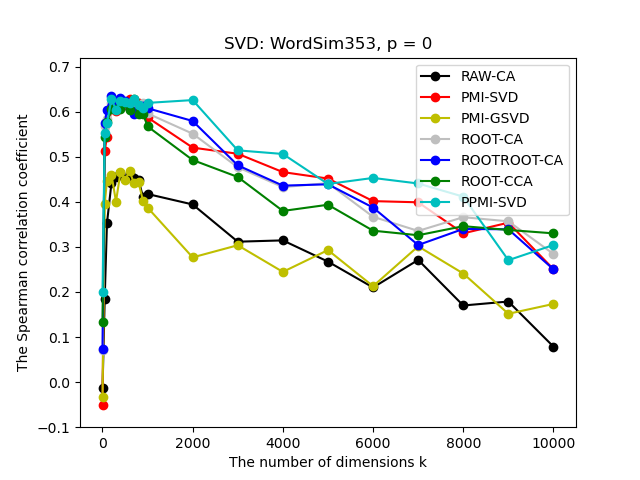}
\caption{WordSim353: $p = 0$}\label{F: bncSVDWordSim3530}
         \end{subfigure}
    \begin{subfigure}[b]{0.42\linewidth}
         \centering
         \includegraphics[width=1\textwidth]{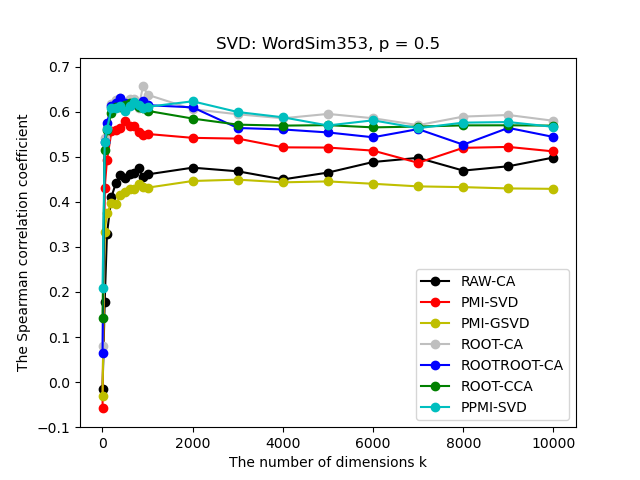}
\caption{WordSim353: $p = 0.5$}\label{F: bncSVDWordSim3530dot5}
         \end{subfigure}
        \begin{subfigure}[b]{0.42\linewidth}
         \centering
         \includegraphics[width=1\textwidth]{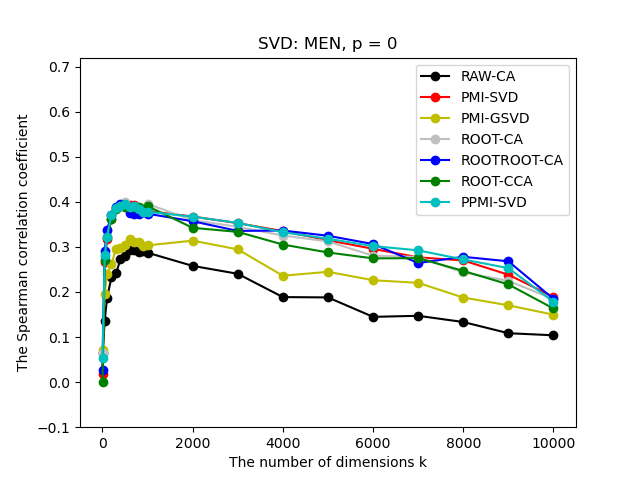}
\caption{MEN: $p = 0$}\label{F: bncSVDMEN0}
         \end{subfigure}
    \begin{subfigure}[b]{0.42\linewidth}
         \centering
         \includegraphics[width=1\textwidth]{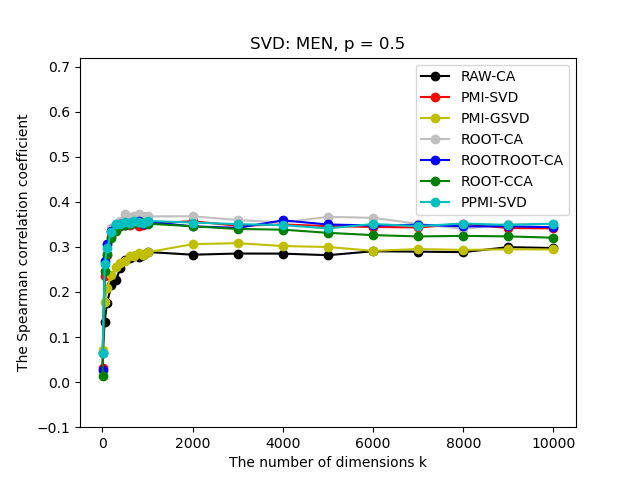}
\caption{MEN: $p = 0.5$}\label{F: bncSVDMEN0dot5}
         \end{subfigure}
        \begin{subfigure}[b]{0.42\linewidth}
         \centering
         \includegraphics[width=1\textwidth]{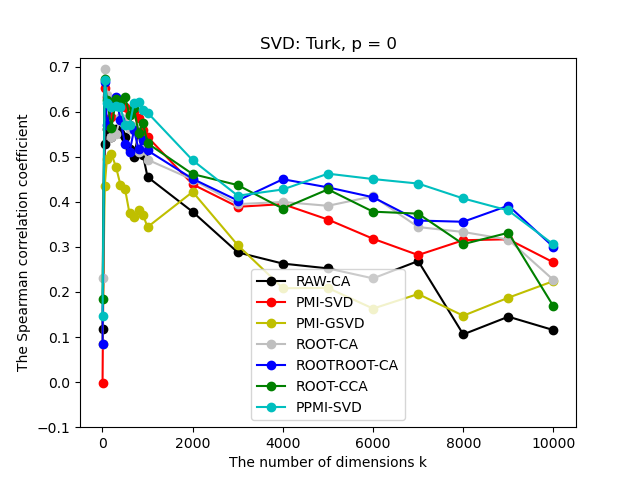}
\caption{Turk: $p = 0$}\label{F: bncSVDTurk0}
         \end{subfigure}
    \begin{subfigure}[b]{0.42\linewidth}
         \centering
         \includegraphics[width=1\textwidth]{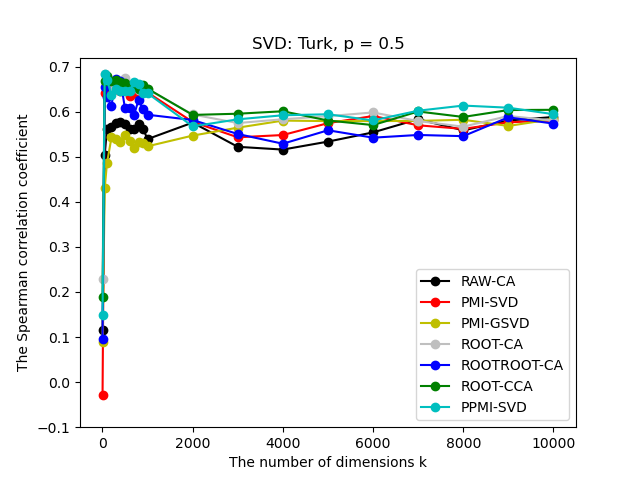}
\caption{Turk: $p = 0.5$}\label{F: bncSVDTurk0dot5}
         \end{subfigure}
        \caption{BNC}
    \label{F: bncSVD1}
\end{figure}
         
\begin{figure}[H]
 \ContinuedFloat
    \centering        
        \begin{subfigure}[b]{0.42\linewidth}
         \centering
         \includegraphics[width=1\textwidth]{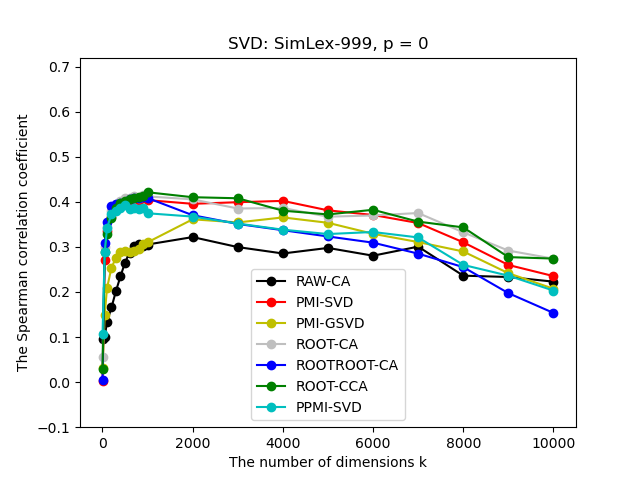}
\caption{SimLex-999: $p = 0$}\label{F: bncSVDSimLex-9990}
         \end{subfigure}
    \begin{subfigure}[b]{0.42\linewidth}
         \centering
         \includegraphics[width=1\textwidth]{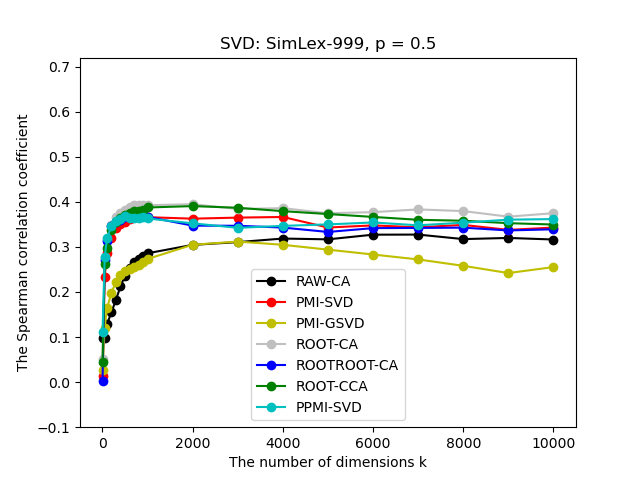}
\caption{SimLex-999: $p = 0.5$}\label{F: bncSVDSimLex-9990dot5}
         \end{subfigure}
         \caption{BNC}
    \label{F: bncSVD2}
\end{figure}

\begin{figure}[H]
    \centering
        \begin{subfigure}[b]{0.42\linewidth}
         \centering
         \includegraphics[width=1\textwidth]{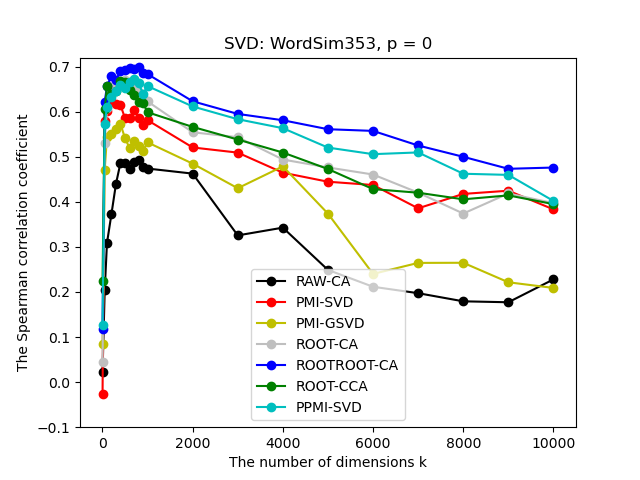}
\caption{WordSim353: $p = 0$}\label{F: wikiSVDWordSim3530}
         \end{subfigure}
    \begin{subfigure}[b]{0.42\linewidth}
         \centering
         \includegraphics[width=1\textwidth]{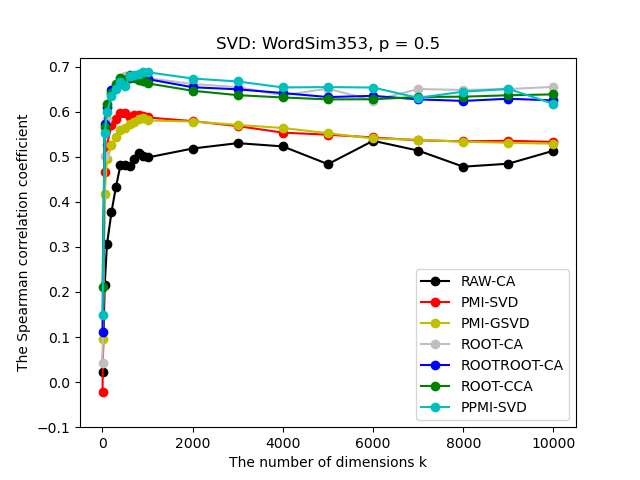}
\caption{WordSim353: $p = 0.5$}\label{F: wikiSVDWordSim3530dot5}
         \end{subfigure}
        \begin{subfigure}[b]{0.42\linewidth}
         \centering
         \includegraphics[width=1\textwidth]{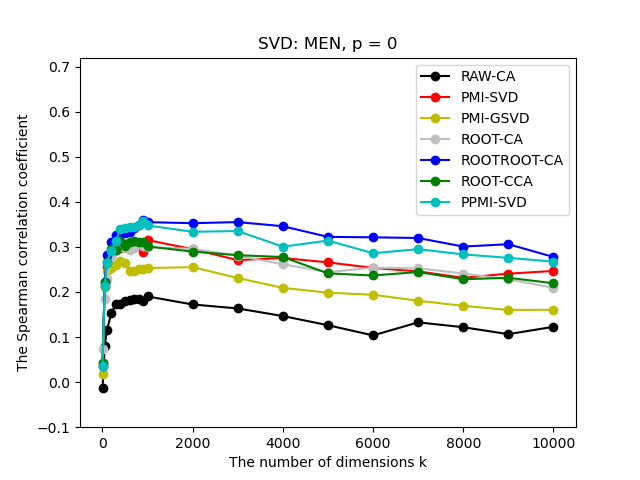}
\caption{MEN: $p = 0$}\label{F: wikiSVDMEN0}
         \end{subfigure}
    \begin{subfigure}[b]{0.42\linewidth}
         \centering
         \includegraphics[width=1\textwidth]{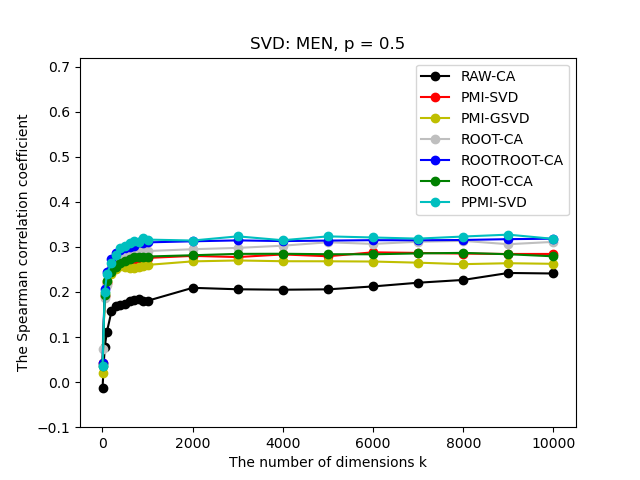}
\caption{MEN: $p = 0.5$}\label{F: wikiSVDMEN0dot5}
         \end{subfigure}
        \begin{subfigure}[b]{0.42\linewidth}
         \centering
         \includegraphics[width=1\textwidth]{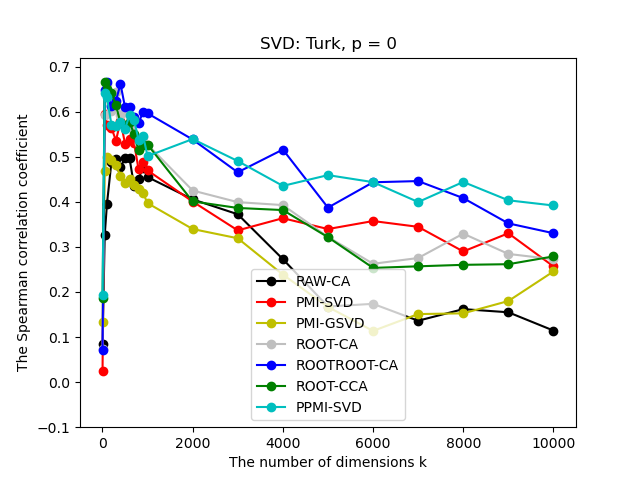}
\caption{Turk: $p = 0$}\label{F: wikiSVDTurk0}
         \end{subfigure}
    \begin{subfigure}[b]{0.42\linewidth}
         \centering
         \includegraphics[width=1\textwidth]{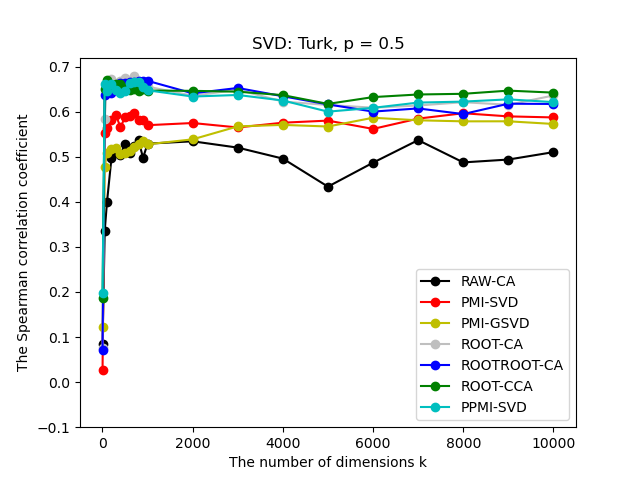}
\caption{Turk: $p = 0.5$}\label{F: wikiSVDTurk0dot5}
         \end{subfigure}
        \caption{Wiki052024}
    \label{F: wikiSVD1}
\end{figure}
         
\begin{figure}[H]
 \ContinuedFloat
    \centering        
        \begin{subfigure}[b]{0.42\linewidth}
         \centering
         \includegraphics[width=1\textwidth]{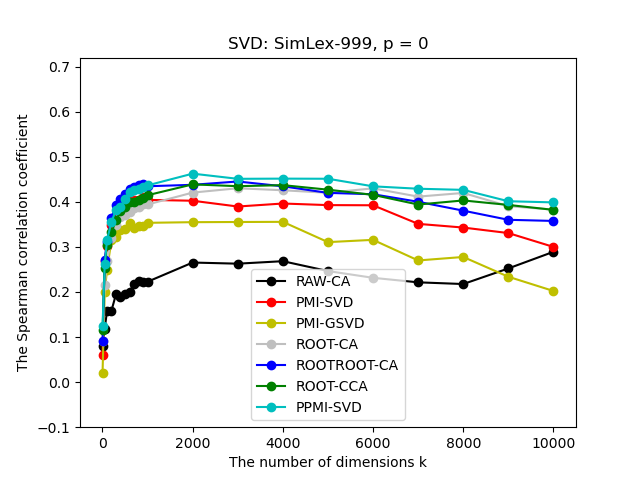}
\caption{SimLex-999: $p = 0$}\label{F: wikiSVDSimLex-9990}
         \end{subfigure}
    \begin{subfigure}[b]{0.42\linewidth}
         \centering
         \includegraphics[width=1\textwidth]{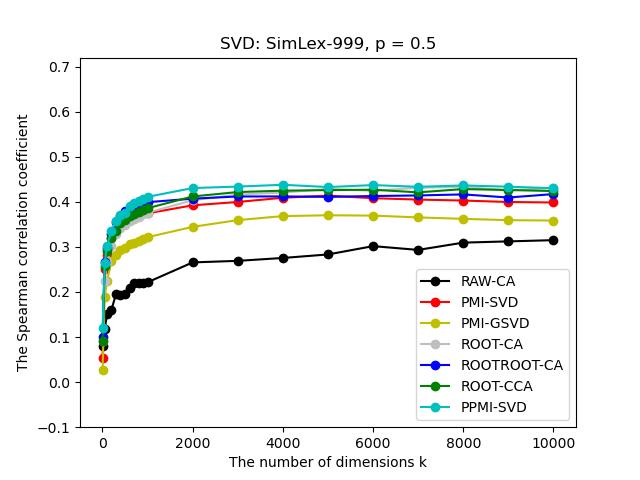}
\caption{SimLex-999: $p = 0.5$}\label{F: wikiSVDSimLex-9990dot5}
         \end{subfigure}
         \caption{Wiki052024}
    \label{F: wikiSVD2}
\end{figure}

\appendixqqsection{BNC: the number and sizes of extreme values of PMI, PPMI, and WPMI, and plots showing the contribution of the rows about PMI-SVD, PPMI-SVD, and PMI-GSVD}\label{BNC: numext}

\begin{figure}
    \centering
    \includegraphics[width=1\textwidth]{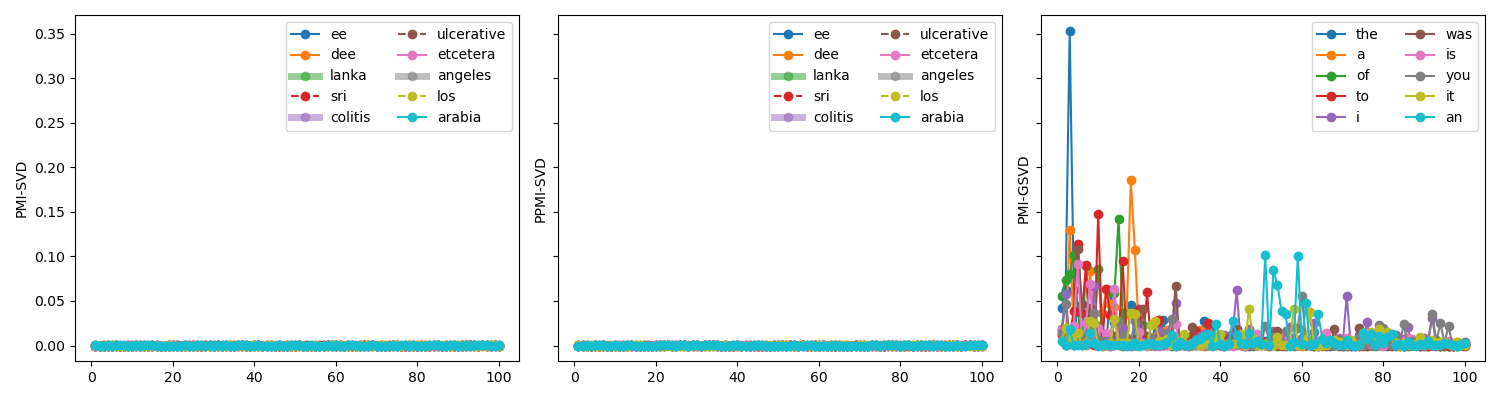}
         \caption{BNC: the contribution of the rows, corresponding to top 10 extreme values, to first 100 dimensions of PMI-SVD, PPMI-SVD, PMI-GSVD.}
    \label{F: bncpmiextremevaluecontri}
\end{figure}

Table~\ref{T: bncnumext}, part PMI, PPMI, WPMI, shows the number of extreme values of PMI, PPMI, WPMI matrices. We only include non-zero pairs of PMI matrix because the PMI matrix is sparse: 84.1\% of the entries are
zero. The corresponding 84.1\% of entries in PPMI and WPMI are also ignored. The number of extreme values of WPMI matrix (2,525,345) is much larger than that of PMI and PPMI (141,366 and 405,830). 
Furthermore, in WPMI the extremeness of the extreme values is much larger than those in PMI and PPMI. For example, where the average contribution of each cell is $1/\left(11,332\times 11,332\right)$, in WPMI the most extreme entry (the, the) contributes around 0.01150 to the total inertia,
while in PMI (PPMI), the most extreme entry (ee, ee) contributes around $2.2 \times 10^{-6}$ ($2.7 \times 10^{-6}$) to the total inertia. Figure~\ref{F: bncpmiextremevaluecontri} shows the contribution of the rows, corresponding to top 10 extreme values, to the first 100 dimensions of PMI-SVD, PPMI-SVD, PMI-GSVD. The rows, corresponding to the top extreme values of WPMI, take up a bigger contribution to the first dimensions of PMI-GSVD. For example, in PMI-GSVD, the “the” row contributes more than 0.3 to the third dimension, while in PMI-SVD and PPMI-SVD, the contributions are much smaller.

\begin{table}[H]
    \centering
        \caption{BNC: the number of extreme values}
    \label{T: bncnumext}
    \begin{tabular}{lllllllllll}
    \hline
        &  LT$f_1$ & GT$f_3$ & total \\
    \hline
PMI           &    13,982 &   127,384 &   141,366 \\
PPMI          &        0 &   405,830 &   405,830 \\
WPMI          &  2,037,800 &   487,545 &  2,525,345 \\
\hline
TTEST      &   334,512 &  1,480,336 &  1,814,848 \\
ROOT-TTEST     &    35,418 &   927,470 &   962,888 \\
ROOTROOT-TTEST &    31,234 &   750,433 &   781,667 \\
STRATOS-TTEST  &        0 &  1,173,717 &  1,173,717 \\
    \hline
    \end{tabular}
\end{table}

\appendixqqsection{Wiki052024: the number and sizes of extreme values of PMI, PPMI, and WPMI, and plots showing the contribution of the rows about PMI-SVD, PPMI-SVD, and PMI-GSVD}\label{Wiki052024: numext}

\begin{figure}
    \centering
    \includegraphics[width=1\textwidth]{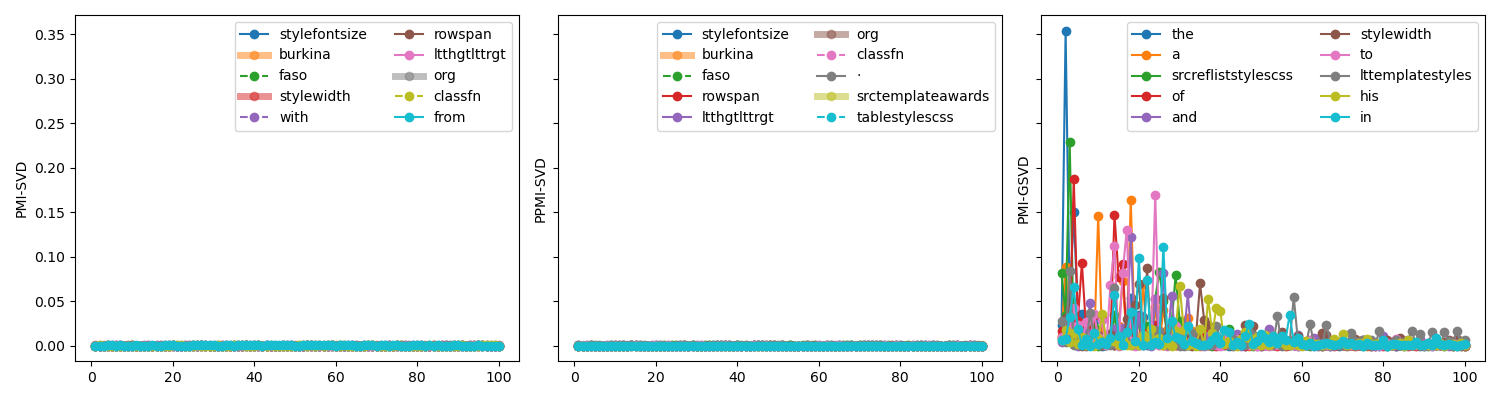}
         \caption{Wiki052024: the contribution of the rows, corresponding to top 10 extreme values, to first 100 dimensions of PMI-SVD, PPMI-SVD, PMI-GSVD.}
    \label{F: wikipmiextremevaluecontri}
\end{figure}

Table~\ref{T: wikinumext}, part PMI, PPMI, WPMI, shows the number of extreme values of PMI, PPMI, WPMI matrices. We only include non-zero pairs of PMI matrix because the PMI matrix is sparse: 56.8\% of the entries are
zero. The corresponding 56.8\% of entries in PPMI and WPMI are also ignored. The number of extreme values of WPMI matrix (9,823,926) is much larger than that of PMI (1,449,329). In this corpus, PPMI also has a large number of extreme values.
Furthermore, in WPMI the extremeness of the extreme values is much larger than those in PMI and PPMI. For example, where the average contribution of each cell is $1/\left(15,135\times 15,135\right)$, in WPMI the most extreme entry (the, the) contributes around 0.01786 to the total inertia,
while in PMI (PPMI), the most extreme entry (stylefontsize, stylefontsize) contributes around $6.4 \times 10^{-7}$ ($2.2 \times 10^{-6}$) to the total inertia. Figure~\ref{F: wikipmiextremevaluecontri} shows the contribution of the rows, corresponding to top 10 extreme values, to the first 100 dimensions of PMI-SVD, PPMI-SVD, PMI-GSVD. The rows, corresponding to the top extreme values of WPMI, take up a bigger contribution to the first dimensions of PMI-GSVD. For example, in PMI-GSVD, the “the” row contributes more than 0.3 to the second dimension, while in PMI-SVD and PPMI-SVD, the contributions are much smaller.

\begin{table}[H]
    \centering
        \caption{Wiki052024: the number of extreme values}
    \label{T: wikinumext}
    \begin{tabular}{lllllllllll}
    \hline
        &  LT$f_1$ & GT$f_3$ & total \\
    \hline
PMI           &   360,804 &   108,8525 &   1,449,329 \\
PPMI          &        0 &  13,333,552 &  13,333,552 \\
WPMI          &  8,329,451 &   1,494,475 &   9,823,926 \\
\hline
TTEST      &  4,829,690 &   7,961,062 &  12,790,752 \\
ROOT-TTEST     &   404,134 &   5,524,112 &   5,928,246 \\
ROOTROOT-TTEST &   140,723 &   2,787,901 &   2,928,624 \\
STRATOS-TTEST  &        0 &   7,642,196 &   7,642,196 \\
    \hline
    \end{tabular}
\end{table}

\appendixqqsection{Text8: plots showing the contribution of the rows about ROOT-CCA}\label{Text8: text8rootccaextremevaluecontri}

\begin{figure}
    \centering
    \includegraphics[width=0.40\textwidth]{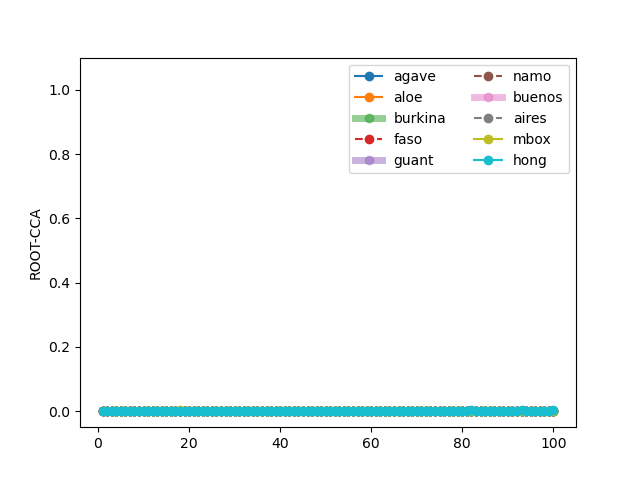}
         \caption{Text8: the contribution of the rows, corresponding to top 10 extreme values, to first 100 dimensions of ROOT-CCA.}
    \label{F: text8rootccaextremevaluecontri}
\end{figure}

Figure~\ref{F: text8rootccaextremevaluecontri} shows the contribution of the rows, corresponding to top 10 extreme values, to first 100 dimensions of ROOT-CCA.

\appendixqqsection{BNC: the number and sizes of extreme values of TTEST, ROOT-TTEST, ROOTROOT-TTEST, and STRATOS-TTEST, and plots showing the contribution of the rows about RAW-CA, ROOT-CA, ROOTROOT-CA, and ROOT-CCA}\label{bnc: bncrootccaextremevaluecontri}

\begin{figure}
    \centering
    \includegraphics[width=1\textwidth]{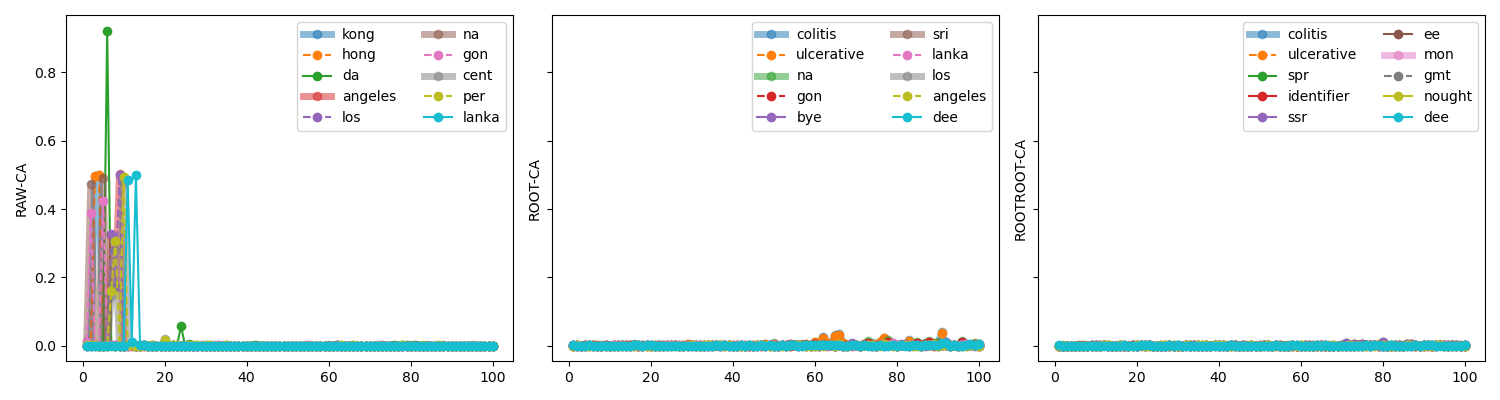}
         \caption{BNC: the contribution of the rows, corresponding to top 10 extreme values, to first 100 dimensions of RAW-CA, ROOT-CA, ROOTROOT-CA.}
    \label{F: bnccaextremevaluecontri}
\end{figure}

\begin{figure}
    \centering
    \includegraphics[width=0.40\textwidth]{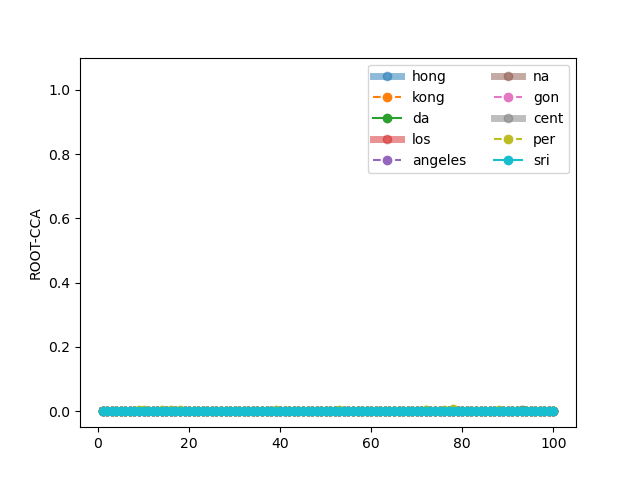}
         \caption{BNC: the contribution of the rows, corresponding to top 10 extreme values, to first 100 dimensions of ROOT-CCA.}
    \label{F: bncrootccaextremevaluecontri}
\end{figure}

The bottom part of Table~\ref{T: bncnumext} shows the number of extreme values of TTEST, ROOT-TTEST, ROOTROOT-TTEST and STRATOS-TTEST matrices. Similar with PMI, PPMI, WPMI, 84.1\% of entries are ignored. The number of extreme values of TTEST matrix (1,814,848) is much larger than that of ROOT-TTEST, ROOTROOT-TTEST and STRATOS-TTEST (962,888, 781,667, and 1,173,717). Furthermore, in TTEST the extremeness of the extreme values is much larger than in ROOT-TTEST, ROOTROOT-TTEST and STRATOS-TTEST. For example, in TTEST the most extreme entry (kong, hong) or (hong, kong) contributes around 0.00965 to the total inertia, while in ROOT-TTEST, ROOTROOT-TTEST and STRATOS-TTEST, the most extreme entries (colitis, ulcerative) or (ulcerative, colitis), (colitis, ulcerative) or (colitis, ulcerative), (hong, kong) or (kong, hong) contribute around 0.00047, 0.00003, and 0.00008 respectively. Figure~\ref{F: bnccaextremevaluecontri} shows the contribution of the rows, corresponding to top 10 extreme values, to first 100 dimensions of RAW-CA, ROOT-CA, and ROOTROOT-CA. The corresponding plot about ROOT-CCA is in Figure~\ref{F: bncrootccaextremevaluecontri}. In RAW-CA, the rows, corresponding to top extreme values of
TTEST, have a big contribution to the first dimensions of RAW-CA, while in ROOT-CA and ROOTROOT-CA, the contributions are much smaller, which also holds for ROOT-CCA.

\appendixqqsection{Wiki052024: the number and sizes of extreme values of TTEST, ROOT-TTEST, ROOTROOT-TTEST, and STRATOS-TTEST, and plots showing the contribution of the rows about RAW-CA, ROOT-CA, ROOTROOT-CA, and ROOT-CCA}\label{wiki: wikirootccaextremevaluecontri}

\begin{figure}
    \centering
    \includegraphics[width=1\textwidth]{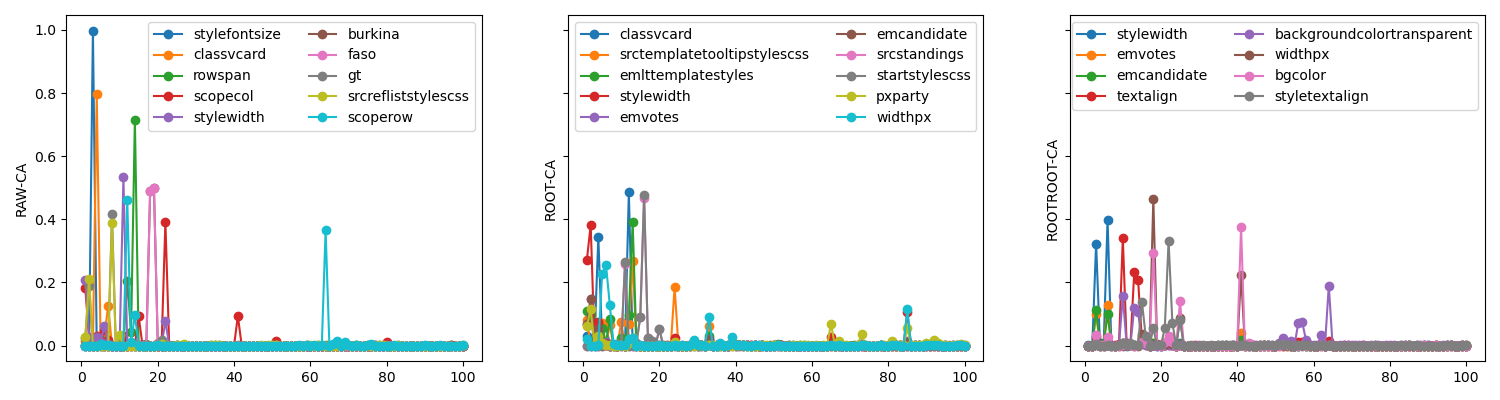}
         \caption{Wiki052024: the contribution of the rows, corresponding to top 10 extreme values, to first 100 dimensions of RAW-CA, ROOT-CA, ROOTROOT-CA.}
    \label{F: wikicaextremevaluecontri}
\end{figure}

\begin{figure}
    \centering
    \includegraphics[width=0.40\textwidth]{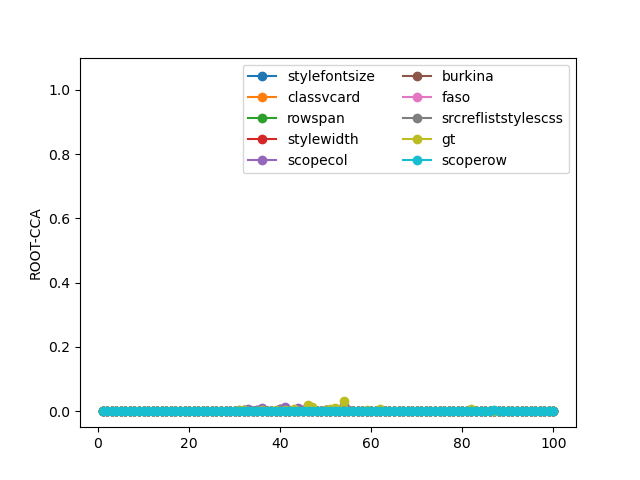}
         \caption{Wiki052024: the contribution of the rows, corresponding to top 10 extreme values, to first 100 dimensions of ROOT-CCA.}
    \label{F: wikirootccaextremevaluecontri}
\end{figure}

The bottom part of Table~\ref{T: wikinumext} shows the number of extreme values of TTEST, ROOT-TTEST, ROOTROOT-TTEST and STRATOS-TTEST matrices. Similar with PMI, PPMI, WPMI, 56.8\% of entries are ignored. The number of extreme values of TTEST matrix (12,790,752) is much larger than that of ROOT-TTEST, ROOTROOT-TTEST and STRATOS-TTEST (5,928,246, 2,928,624, and 7,642,196). Furthermore, in TTEST the extremeness of the extreme values is much larger than in ROOTROOT-TTEST and STRATOS-TTEST. For example, in TTEST the most extreme entry (stylefontsize, stylefontsize) contributes around 0.01359 to the total inertia, while in ROOTROOT-TTEST and STRATOS-TTEST, the most extreme entries (stylewidth, emvotes) or (emvotes, stylewidth), (stylefontsize, stylefontsize) contribute around 0.00448 and 0.00012 respectively. In ROOT-TTEST, the most extreme entry (classvcard, classvcard) contributes around 0.01825, which is larger than that in TTEST 0.01359. Figure~\ref{F: wikicaextremevaluecontri} shows the contribution of the rows, corresponding to top 10 extreme values, to first 100 dimensions of RAW-CA, ROOT-CA, and ROOTROOT-CA. The corresponding plot about ROOT-CCA is in Figure~\ref{F: wikirootccaextremevaluecontri}. In RAW-CA, the rows, corresponding to top extreme values of
TTEST, have a big contribution to the first dimensions of RAW-CA, while in ROOT-CA and ROOTROOT-CA, the contributions are much smaller, which also holds for ROOT-CCA. For example, in RAW-CA, the “Stylefontsize” row contributes around 0.998 (almost 100 percent) to the third dimension.

\end{document}